\documentclass[journal]{IEEEtran}
\usepackage{amsmath,amssymb}
\usepackage{booktabs}
\usepackage{graphicx}
\usepackage[hidelinks]{hyperref}
\usepackage{multirow}
\usepackage{xcolor}

\newcommand{\osfurl}{\url{https://osf.io/fspuc}}
\newcommand{\repourl}{\url{https://github.com/maxsegan/moba-draft-policies}}
\newcommand{\deployurl}{\url{https://www.hotsfever.com}}

\begin{document}

\title{Searching Beyond the Data: Value-Function Exploitation and the Cost of Staying on Support in MOBA Drafting}
\author{Max~Segan, Ernest~Pysher, James~Frankel, and Hod~Lipson%
\thanks{M. Segan is with the Department of Computer Science and H. Lipson with the Department of Mechanical Engineering, Columbia University (e-mail: m.segan@columbia.edu; hod.lipson@columbia.edu). E. Pysher and J. Frankel are independent researchers (e-mail: ernestpysheriv@gmail.com; james@frankel.tv).}}

\maketitle

\begin{abstract}
Search over a learned value function can find the value function's mistakes in states the data barely cover. We measure this in drafting for Heroes of the Storm, using 1.94 million logged drafts. Monte Carlo tree search over a learned win-probability model does leave the data: 63\% of its teams contain a three-hero combination that humans played together fewer than 25 times, against 5\% of human teams. Yet on 260,539 later real games, the model is no more wrong on rare teams than on common ones; its main measured error is a winner's curse of about 1.4 points on the teams it likes most. In simulated worlds fit to the data, where every team's true value is known, the value function's errors cost the search 1.6 points, while blocking rare teams at play time costs 2.2 to 2.7. Training the search's prior under the same floor cuts the floor's cost from 2.6 points to 1.1 over eight training seeds. With prior and floor together, 99.7\% of teams stay on support with no detected loss against the unconstrained policy. The pessimistic offline reinforcement learning methods we tested either lose strength or leave the data more than search does.
\end{abstract}

\begin{IEEEkeywords}
MOBA drafting, Monte Carlo tree search, offline reinforcement learning, distribution shift, value-function exploitation.
\end{IEEEkeywords}

\section{Introduction}

AlphaZero-style search turns a value function into a decision-maker. It looks ahead, scores the positions it reaches, and picks the move whose consequences score best~\cite{silver2018general}. When the value function is learned from logged human play, the method carries a known risk. The search is an optimizer, and an optimizer seeks out wherever its objective is too high. If the value function is most wrong where the data are thin, search will go there. The same concern drives pessimism in offline reinforcement learning~\cite{kumar2020conservative,levine2020offline} and the study of reward-model overoptimization~\cite{gao2023scaling}.

The usual responses are to penalize unfamiliar actions or to keep the policy near the data. Both cost something. A drafter kept near human play inherits human habit, and a penalty large enough to stop exploitation can stop discovery too. Whether that cost is worth paying depends on facts that are rarely measured: how far search actually goes, whether the value function is in fact more wrong out there, and what staying close would give up.

Drafting in multiplayer online battle arena (MOBA) games lets us measure all three. In Heroes of the Storm, two teams of five alternate through 16 actions, banning 6 heroes and picking 10 from a pool of 90, before the match is played. We have 1.94 million logged ranked drafts on one fixed roster. A value function that sees only the draft predicts the match at about 57\%, a large edge in a balanced game. Search can reach almost any team: there are about $7.2 \times 10^{14}$ possible final matchups, and our data cover a small and clustered fraction of them.

We ask four questions and answer each with measurements.

\textbf{Does search leave the data?} Yes. With a role-feasibility mask, 63\% of the MCTS drafter's teams contain a three-hero subset that humans fielded together fewer than 25 times in training. Without the mask the share is 72\%. For human teams it is about 5\%.

\textbf{Is the value function more wrong out there?} Not measurably, on real games. A calibration slope on the model's claimed team-interaction value, 1 when the claims are calibrated, is 0.92 on rare teams and 0.87 on common ones. Seven rarity measures, a backtest on models trained on older data, and Quick Match, a mode without a draft, all agree. What the value function does get wrong is the human teams it likes most: the 10\% with the largest claimed interaction value are overvalued by about 1.45 points, rare or common alike. This is a winner's curse, and a support floor does not target it.

\textbf{What do its errors cost the search?} We build simulated worlds from an explicit Bayesian model of the same games, so that every team, seen or not, has a known true value. Replacing the search's value function with the truth gains it 1.6 points on average (95\% interval 0.2 to 3.6). Refitting the value function inside a world, from outcomes drawn there, shows no systematic overclaim by the search ($-$0.6 points, $-$1.8 to +0.2).

\textbf{What does staying on support cost, and can we avoid paying it?} Forbidding moves that create a rare triple costs 2.2 to 2.7 true points in most settings. In the main world family it pays for itself only if every rare team is about 4.5 points worse than the model believes; on real human teams the 95\% bound is about 2. Training the search's prior under the same floor cuts the floor's cost from 2.6 points to 1.1 over eight training seeds of each policy, and every floor-trained seed pays less than every ordinary one. The floor-trained prior is also stronger without the floor, by 1.6 points (0.4 to 2.8). With the floor and decision-time search it is +0.8 points ($-$0.5 to 2.0) against the unconstrained MCTS policy, with 0.3\% of teams below the floor, and +2.0 (0.7 to 3.3) if every rare team is 2 points worse than the model believes.

Our contributions:
\begin{enumerate}
 \item A measurement of how far search over a learned value function leaves its data, on seven support measures, for MCTS, greedy value search, behavior-anchored methods, and offline RL.
 \item Real-data evidence on whether a learned value function is more wrong off support, from 260,539 games no agent trained on, a temporal backtest, and an undrafted game mode. It is not, and the error that remains is a winner's curse that does not depend on support.
 \item Ground-truth worlds fit to the same data, with dials for how wrong rare teams might be, used to measure what the value function's errors cost the search and what a support floor costs it.
 \item A drafter that stays on support at no measured cost: a floor at play time on a prior trained under the same floor, replicated over eight training seeds. Pessimistic offline RL (CQL, IQL, MCQ, and a thresholded penalty) does not reach the same point.
\end{enumerate}
Code, models, and a fused CUDA MCTS kernel are open source,\footnote{\repourl} and the full agent runs in the browser.\footnote{\deployurl}

\section{Background and Related Work}

\textit{Search over learned values.} Search amplifies whatever errors its value function has~\cite{silver2016mastering,silver2018general}. The effect is well documented for reward models under policy optimization, where the proxy reward rises while the true reward falls~\cite{gao2023scaling}. In board-game search the final evaluation is exact. In drafting it must be learned from the same data the policy needs to move beyond.

\textit{Offline RL and pessimism.} Offline RL learns from a fixed dataset and guards against overestimating actions the data do not contain~\cite{levine2020offline}. Conservative Q-Learning (CQL)~\cite{kumar2020conservative} pushes down the values of unlogged actions. Implicit Q-Learning (IQL)~\cite{kostrikov2022iql} never queries unseen actions. Mildly Conservative Q-Learning (MCQ)~\cite{lyu2022mildly} gives unlogged actions pseudo-targets from in-support ones. BEAR~\cite{kumar2019stabilizing} and TD3+BC~\cite{fujimoto2021minimalist} constrain the policy toward the behavior policy. Each bets that the value function is more wrong where the data are thin. We measure that bet directly.

\textit{Support, uncertainty, and the optimizer's curse.} Safe policy improvement with baseline bootstrapping (SPIBB)~\cite{laroche2019spibb} falls back to the behavior policy wherever a state-action count is below a threshold, the same kind of count floor we study, applied to actions rather than to team compositions. Uncertainty-penalized methods subtract an ensemble's disagreement from the value~\cite{yu2020mopo,an2021edac}, and the theory of offline learning ties its guarantees to how well the data cover the states and actions the learned policy visits~\cite{rashidinejad2021bridging}. All of these treat thin data as the source of overestimation. Decision analysis names a second source that needs no thin data: choosing the option with the highest estimate selects for estimates that are too high, the optimizer's curse~\cite{smith2006optimizer}. We measure both in one setting. For this value function, the second is the one we can detect.

\textit{Drafting.} MOBA agents such as OpenAI Five~\cite{berner2019dota} focus on in-match control. Draft work either predicts outcomes and optionally searches against the predictor~\cite{chen2018drafting,chen2021juewudraft}, or imitates expert selections~\cite{summerville2016draft,gourdeau2021discriminative,lee2022draftrec}. Gourdeau and Archambault~\cite{gourdeau2021discriminative} conjectured that search over a win-rate estimator would wander into poorly characterized compositions, and proposed constraining it to the professional manifold. DraftRec~\cite{lee2022draftrec} restricts recommendations to observed combinations. We test the conjecture and price the constraint.

\section{Setup}
\label{sec:setup}

\subsection{Data and Value Function}

A draft is 16 actions (6 bans, 10 picks) in a fixed order. The state holds each team's picks and the bans, the map (14), the skill tier (3), and the step. The reward is the game outcome. Our corpus is 1,943,338 ranked Storm League replays on patch 2.55, December 2021 to May 2026, with the hero pool fixed at 90. Tiers come from each game's average rank: low is Bronze and Silver (41\% of games), mid is Gold and Platinum (41\%), high is Diamond and above (18\%). Splits are at the replay level: 1,866,369 replays train, 38,100 validate, and 38,869 are held out for test. A further 260,539 games uploaded after the snapshot, on the same balance build, were used by no agent, value function, or agent statistic. We call them the post-snapshot games. They carry the main real-data test of Section~\ref{sec:wrong}, and the judges of Section~\ref{sec:refs} are trained on them.

The value function is a win-probability (WP) network over hero identities, map, tier, and 86 features built from training statistics: role counts, per-hero counter and synergy deltas, hero and map win rates, and the empirical win rate of each team's role composition. A statistic that includes a game's own outcome leaks the label into that game's features, so every statistic is computed out of fold: each training row's features come from the other four of five hash folds, and every other row uses statistics over the whole training split. The value function reaches 57.5\% on test and 56.8\% on post-snapshot games, with calibration slope 0.96 on the latter. With in-sample statistics the same model is overconfident (slope 0.75), and search against it raises its own score while independent judges fall (supplement).

\textit{Data availability.} Raw replays cannot be redistributed under the Heroes Profile terms of service. The repository includes the replay-id list, the splits and folds, and the code that rebuilds every statistic.

\subsection{Search}
\label{sec:mcts}

The drafter is an AlphaZero-style policy-value network (4.4M parameters). At each of its own picks and bans the search selects a child by
\begin{equation}
a^* = \arg\max_a \; Q(s,a) + c\,P(s,a)\,\frac{\sqrt{N(s)}}{1+N(s,a)},
\end{equation}
with prior $P$ from the policy network, visit counts $N$, and $c = 2$. Opponent turns are chance nodes. Their replies come from a behavioral-cloning drafter (GD), and children are added by progressive widening (at most $\lceil\sqrt{N}\rceil$ after $N$ visits). A leaf is completed to a full draft by a GD rollout and scored by the WP model, averaged over the two seat orders. Every backed-up value is the WP model's score of a finished draft, so the search optimizes that model. The policy is trained by MCTS against a pool of five GD opponents. Training runs in a fused CUDA kernel that keeps the whole episode, tree, and networks on the GPU (63 episodes per second at 200 simulations per move on one RTX 5090; supplement). The operating point trains at 400 simulations per move for 300K episodes, and we pool five seeds. The trained policy can play directly, or serve as the prior of a fresh search at decision time (200 simulations).

Every search drafter also uses a \textit{role mask}. A pick is removed if it would leave no completion with a healer, a frontline hero, and ranged damage, and at most two heroes of a stacking role. The mask is decided exactly by enumeration. It removes teams that lack a required role, so the results below concern rarity. The mask is not free: in the simulated worlds of Section~\ref{sec:worlds}, MCTS without it is 1.7 true points stronger (0.2 to 3.0). The world model has no role term, so this may understate what role-broken teams lose in real games. We keep it because a drafter for human players must not field such teams.

\subsection{Baselines}
\label{sec:baselines}

All baselines use the same training split. \textit{Greedy value search} completes each candidate pick with GD rollouts and takes the best by the WP model. \textit{Highest win rate} takes the available hero with the best win rate on the map and tier. The offline RL methods are trained on the logged drafts with the game result as the reward. CQL is swept over its penalty weight (0.1 to 5), on hero identities and on the enriched features. IQL is run at five settings of its expectile and temperature (0.7 with 3, 10, or 30; 0.9 with 3 or 30). MCQ is swept over its mixing weight (0.5, 0.7, 0.9). The \textit{thresholded penalty} replaces CQL's penalty with one that pushes down only values above a threshold, in the spirit of MCQ. Table~\ref{tab:frontier} reports one representative setting of each. The anchored methods are well trained by the usual measure: they match the logged human action as often as the behavioral-cloning drafter (top-1 13.3 to 13.5\% against 13.0 to 13.2\%). Across the sweeps, CQL and IQL put 14 to 33\% of their teams below the floor and are judged at 42 to 44\% (supplement).

\subsection{Support Measures and the Floor}
\label{sec:support}

A team is \textit{off support} when the training data say little about it. Our main measure is the count of its rarest three-hero subset among the 3.73 million training teams, the \textit{rarest triple}. Six others check that the results do not depend on that choice: the rarest pair, the rarest four-hero subset, the count of the exact five, the mean pointwise mutual information of the team's ten pairs (low when its heroes are rarely played together), the count of its role composition, and the rarest cross-team matchup pair. For each measure, \textit{below the floor} means below the 5th percentile of held-out human teams. For measures whose 5th percentile is a count of zero, below the floor means a count of zero. For the rarest triple the percentile is 23, which we round to 25 as the operating \textit{floor}. A drafter \textit{uses the floor} when it makes no pick that creates a triple seen fewer than 25 times, unless no legal hero passes; then it takes the hero whose rarest triple is most common. This fallback is why floored search keeps 0.3 to 1\% of teams below the floor (greedy search 3\%, the win-rate drafter 9\%).

\subsection{Judges}
\label{sec:refs}

A drafter cannot be graded by the value function it searched. We score drafts with judges that share no game, statistic, or search with any agent: \textit{gN}, three enriched models trained only on post-snapshot games; \textit{gN-naive}, a hero-identity model on the same games; \textit{RN}, a cross-fitted index of realized hero, pair, and composition win rates; and \textit{QM2026}, a model of Quick Match, a mode with no draft. The consensus is their mean. Each judge is audited on real teams with a missing role, which it would otherwise under-penalize, and corrected with three structure terms (supplement).

The judges are themselves value functions trained on human teams, so they too might misjudge rare teams. On real games they underpredict how badly the rarest 1\% of human teams do, by 4.0 points for the consensus. We therefore report every judged score twice: as judged, and after a support correction fit by cross-validation on 1M training games (supplement). The difference between the two scores is the part due to the judges' support bias. A judged score is the mean win probability against eight shared opponents (MCTS with and without the mask, enriched and masked greedy search, a featureless MCTS agent, the Gourdeau estimator and discriminator, and GD), both seats, 200 drafts per ordered pair.

\textit{Expert judgment.} Every judge here is a model, so we also tested whether expert players could provide a check that is not. In a study registered before data collection (\osfurl), nine Master-rank raters judged 240 items each. On pairs of machine drafts they agreed with the models no more often than chance (0.50, 95\% CI 0.43 to 0.57), and where a behavior-anchored draft met a searching one they preferred the human-like draft. On 570 real games they picked the winner 60\% of the time at the lowest tier and were indistinguishable from chance above it, where draft-only models score 57 to 58\%. The study did not establish expert judgment as an independent check on the models. The evaluation therefore rests on outcomes: held-out real games for the value function's errors, and simulated worlds with a known truth for the search's (supplement).

\section{Search Leaves the Data}
\label{sec:leaves}

\begin{figure}[t]
\centering
\includegraphics[width=\columnwidth]{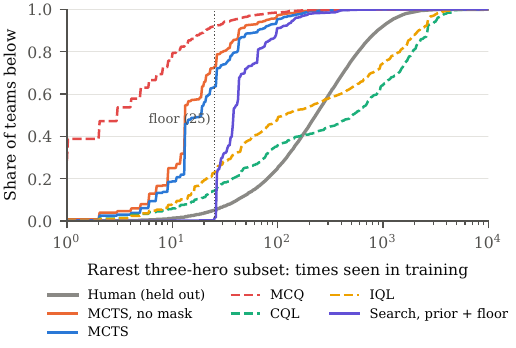}
\caption{Cumulative distribution of each drafter's rarest three-hero subset, counted in training. A curve that rises early means more teams built on rare combinations. The dotted line is the floor. Human teams are a 20,000-team sample from the validation split (median 229; Table~\ref{tab:support} uses post-snapshot human teams, median 215). Drafters' teams come from their games against the eight shared opponents.}
\label{fig:support}
\end{figure}

\begin{table}[t]
\centering
\caption{Support of each drafter's final teams. Below: share with a rarest triple seen fewer than 25 times in training. Median: median rarest-triple count. PMI: mean pair affinity (lower means heroes rarely played together). Exact: share of teams seen exactly in training. Drafters play the eight shared opponents (3,200 or more teams each).}
\label{tab:support}
\setlength{\tabcolsep}{4pt}
\begin{tabular}{lcccc}
\toprule
Drafter & Below & Median & PMI & Exact \\
\midrule
Human (held out) & 5\% & 215 & $-$0.10 & 63\% \\
\midrule
MCTS, no mask & 72\% & 13 & $-$0.38 & 4\% \\
MCTS & 63\% & 19 & $-$0.31 & 5\% \\
Greedy value search & 66\% & 13 & $-$0.21 & 11\% \\
Highest win rate & 81\% & 7 & $-$0.20 & 8\% \\
\midrule
CQL & 14--17\% & 314--513 & $-$0.21 & 62--67\% \\
IQL & 23\% & 114 & $-$0.25 & 57\% \\
MCQ & 92\% & 3 & $-$0.22 & 12\% \\
Thresholded penalty & 66\% & 14 & $-$0.32 & 10\% \\
\bottomrule
\end{tabular}
\end{table}

Search leaves the data, and so does every method that ranks heroes by a learned value (Table~\ref{tab:support}). The masked MCTS drafter puts 63\% of its teams below the floor, with a median rarest triple of 19 against 215 for human teams. Masked greedy search, which completes each candidate draft with behavioral-cloning opponents and picks the best, puts 66\% there. Even a drafter that simply takes the available hero with the highest win rate does: its teams combine strong heroes that people rarely play together.

Fig.~\ref{fig:support} shows the whole distribution. The MCTS curve climbs steeply between 10 and 25: 44\% of its teams have a rarest triple in that range, against 4\% of human teams. The rare region the search visits is close to human play. 70\% of the masked search's below-floor teams have a rarest triple between 10 and 24, and only one of its 17,200 teams contains a triple never seen. Its rare teams sit at the edge of the data.

The search also returns to the same few combinations. Its 10,833 below-floor teams rest on only 164 distinct rarest triples, and the ten most common account for 62\% of them (Table~\ref{tab:triples}, all of the policy's tournament games). One triple, Deathwing with Hogger and Samuro, seen together 13 times in 3.73 million training teams, underlies 34\% of the below-floor teams and 22\% of all the policy's teams. The policy builds around a few rare cores that its value function rates highly. This also limits what the simulated worlds below can say: their verdict on the policy rests largely on a handful of cells, each seen a dozen or so times.

\begin{table}[t]
\centering
\caption{The rarest triples behind the MCTS policy's below-floor teams, over all its tournament games (17,200 teams, 10,833 below the floor). Seen: times the three heroes were on one team in training. Share: share of below-floor teams whose rarest triple it is.}
\label{tab:triples}
\setlength{\tabcolsep}{4pt}
\begin{tabular}{lrr}
\toprule
Rarest triple & Seen & Share \\
\midrule
Deathwing, Hogger, Samuro & 13 & 34.3\% \\
Cho, Samuro, Sgt.\ Hammer & 7 & 5.5\% \\
Artanis, Deathwing, Hogger & 19 & 4.3\% \\
Auriel, Deathwing, Rehgar & 6 & 3.9\% \\
Cho, Hogger, Samuro & 9 & 3.2\% \\
Artanis, Hogger, Samuro & 23 & 2.7\% \\
\midrule
All 164 distinct triples & 0--24 & 100\% \\
\bottomrule
\end{tabular}
\end{table}

Methods that anchor to the logged actions stay closer. CQL and IQL keep 57 to 67\% of their teams exactly as humans have fielded them. They still put 14 to 23\% of teams below the floor, three to five times the human rate, because their departures are scattered rather than aimed. MCQ and a thresholded CQL penalty, whose penalties do not pull toward the logged action, leave the data more than search does. 28\% of MCQ's teams contain a triple never seen in training.

\section{Is the Value Function Wrong off Support?}
\label{sec:wrong}

Leaving the data matters if the value function is more wrong there. We test this on the 260,539 post-snapshot games. For each real team we compare the value function's claim with the outcome, after removing what an additive model of hero strength, map, and tier already explains. The quantity of interest is $\beta$, the share of the value function's claimed non-additive value (its team-interaction value) that shows up in results. Let $b$ be the additive model's win probability for a team, $p$ the value function's, and $y$ the result. The claimed interaction is $I = \operatorname{logit} p - \operatorname{logit} b$, and we fit
\begin{equation}
\Pr(y = 1) = \sigma\left(\operatorname{logit} b + \alpha + \beta I\right)
\end{equation}
separately on teams below and above the floor. $\beta$ is a calibration slope on the claimed interaction log-odds, and $\alpha$ absorbs any level shift. At $\beta = 1$ the interaction claims are right in scale; below 1 they are too large. We read $\beta$ as the share of claimed interaction value that is realized. Standard errors are clustered by game. The main tests keep games whose opponent team is on support, so that only one side's rarity varies.

\begin{figure*}[t]
\centering
\includegraphics[width=\textwidth]{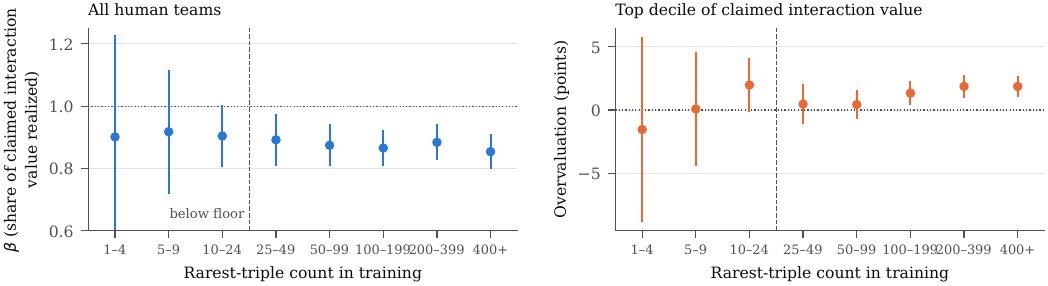}
\caption{The value function's claims against real results on 260,539 post-snapshot games, by the team's rarest-triple count (bins with at least 1,000 teams; opponent on support). Left: $\beta$, the calibration slope on the claimed interaction log-odds (1 = right in scale). Right: overvaluation of the 10\% of teams in each bin with the largest claimed interaction value, in win-probability points (0 = no overvaluation). The dashed line is the floor. Bars: 95\% intervals, clustered by game.}
\label{fig:beta}
\end{figure*}

\begin{table}[t]
\centering
\caption{Share of claimed interaction value that shows up in real results ($\beta$), below and at or above each measure's floor (the 5th percentile of human teams). 260,539 post-snapshot games; $\pm$ one SE clustered by game.}
\label{tab:measures}
\setlength{\tabcolsep}{4pt}
\begin{tabular}{lcccc}
\toprule
Measure & Humans below & $\beta$ below & $\beta$ above & Difference \\
\midrule
Rarest triple & 5.1\% & 0.92 $\pm$ .04 & 0.87 $\pm$ .02 & +0.05 $\pm$ .05 \\
Rarest pair & 5.0\% & 0.95 $\pm$ .05 & 0.87 $\pm$ .02 & +0.08 $\pm$ .05 \\
Rarest four & 7.8\% & 0.91 $\pm$ .04 & 0.87 $\pm$ .02 & +0.04 $\pm$ .04 \\
Exact five & 36.6\% & 0.88 $\pm$ .02 & 0.88 $\pm$ .03 & 0.00 $\pm$ .04 \\
Pair affinity & 5.0\% & 0.93 $\pm$ .05 & 0.85 $\pm$ .02 & +0.08 $\pm$ .05 \\
Role comp. & 5.1\% & 0.92 $\pm$ .04 & 0.87 $\pm$ .02 & +0.05 $\pm$ .05 \\
Matchup & 5.0\% & 0.81 $\pm$ .07 & 0.88 $\pm$ .02 & $-$0.07 $\pm$ .07 \\
\bottomrule
\end{tabular}
\end{table}

\paragraph{Rare human teams are not overrated more} By the rarest triple, $\beta$ is 0.92 below the floor and 0.87 above it (Table~\ref{tab:measures}). On post-snapshot games the value function's interaction claims are about 10\% too large on every kind of human team, and no larger on rare ones. Fig.~\ref{fig:beta} (left) shows no trend with rarity: $\beta$ stays between 0.85 and 0.92 from a rarest-triple count of 1 to above 400. The other six measures agree: the below-floor share runs from 0.81 to 0.95 and is never significantly below the in-support share. The measures are correlated (0.46 to 0.95 with the rarest triple, by percentile), so they are robustness checks rather than independent replications.

\paragraph{Backtest} The most direct test of whether a model is wrong about teams it has not seen is to train it on older data. Models trained on data ending four months and one year before the post-snapshot period were scored on later games (298,607 and 828,886) and on teams that were below the floor in their own training data. Those teams were not overrated. For the four-month model $\beta$ was 1.00 $\pm$ 0.04 below the floor and 0.94 $\pm$ 0.02 in support; for the one-year model, 1.08 $\pm$ 0.02 against 1.00 $\pm$ 0.01. We report the pooled split and the teams that stayed rare. Teams that later became common did much better than predicted, but players chose to adopt them, so that split is confounded.

\paragraph{Quick Match} Quick Match teams are not drafted: each player picks a hero before matchmaking. On 101,549 Quick Match games, compositions are much wider, with 23.6\% of teams below the floor against 5.4\% in ranked play. The slope is again flat across support (0.88 $\pm$ 0.03 below, 0.84 $\pm$ 0.03 above). Below-floor teams there carry a level overvaluation of 0.98 $\pm$ 0.26 points, within an overall shift of domain (calibration slope 0.81 on Quick Match).

\paragraph{The error that remains is a winner's curse} Search does not choose typical teams. It chooses the teams its value function likes most. Among the 10\% of teams with the largest claimed interaction value, real results fall short of the claim by 1.45 $\pm$ 0.95 points below the floor and by 1.44 $\pm$ 0.23 points in support. The model's top-decile teams win about 1.4 points less often than predicted, on and off support, so a support floor does not target this error. By bin (Fig.~\ref{fig:beta}, right), the overvaluation is positive in every bin from 10 upward, 0.4 to 2.0 points, and it is no larger in the rare bins, where the intervals are wide. The search's own teams are less extreme: they claim about 2 points of interaction value against 10 for the human top decile, and the fit implies they are not overvalued ($-$0.4 to $-$0.6 points), in line with the retrained-world result of Section~\ref{sec:worlds}. The size of the curse depends on the model and the baseline. Against a hero-strength baseline it is 3.2 to 3.8 points. The backtest models show less of it: their in-support top decile is overvalued by 0.7 points (four-month model) and 0.0 (one-year), and their below-floor top decile is not overvalued.

\paragraph{What these tests cannot show} Every team in these tests was chosen by human players. A search-chosen team at the same rarity may be less well determined by the data: in the simulated worlds below, posterior 90\% intervals cover the truth for 89 to 90\% of human teams but only 63 to 89\% of search-chosen ones. The 95\% upper bound on extra overvaluation below the floor is about 2 points, close to what the floor costs (Section~\ref{sec:floor}). Real data alone cannot settle whether the floor pays, so we answer it in the worlds.

\section{Ground-Truth Worlds}
\label{sec:worlds}

\subsection{Construction}

We fit an explicit Bayesian outcome model to the training games. A team $T$ on map $m$ at tier $t$ has strength
\begin{multline}
s(T) = \sum_{h \in T} \left(w_h + V_{mh} + U_{th}\right) + \sum_{\{h,h'\} \subset T} S_{hh'} \\
+ \sum_{\{h,h',h''\} \subset T} \langle e_h, e_{h'}, e_{h''} \rangle,
\end{multline}
with per-hero strength $w$, map and tier terms $V$ and $U$, pair synergy $S$, and a rank-8 trilinear term for three-hero interactions. The log-odds that team $A$ beats team $B$ is $s(A) - s(B) + \sum_{a \in A, b \in B} K_{ab}$, with antisymmetric pair counters $K$. Priors are Gaussian, with scales chosen on the validation split, and the posterior is approximated around its mode. It matches the WP model's held-out log loss (0.6773 against 0.6782 on post-snapshot games), and its pair effects have the spread of the real ones (synergy SD 1.78 points against 1.72 measured directly; counters 1.06 against 1.09). A \textit{world} is one draw from the model's posterior, plus independent effects for higher-order cells (triples, quartets, fives, cross-team pair-hero cells, hero-map-tier cells), so that every team, seen or not, has a fixed true value. Effects for cells the data never show are drawn from the prior. Six families each put their uncertainty in one place: three-hero cells (family a, the main family), cross-team cells (b), four- and five-hero cells (c), hero-map-tier cells (d), a balance-patch drift of the size measured between builds (e), and worlds drawn from the drafter's own network class (g).

Four dials, fixed before any agent was scored, set how wrong the model might be off support, scaled by how far below the floor a team is: a level shift $\delta$ in rare teams' strength, a scale $a$ on their interaction value, extra team-level noise $\sigma$, and a scale $k$ on the spread of higher-order cells. The defaults come from the real-data measurements of Section~\ref{sec:wrong}. Named settings move them outside the real-data bounds. \textit{Certify14} sets $\delta = -2.7$ points at full grade, which makes the observed dip of $-$1.8 points at rarest triple 1 to 4 typical rather than extreme (worlds average $-$1.0; the observed value is at their 23rd percentile). \textit{Stress} sets $k = 2$, $\delta = -3$, $\sigma = 3$, and doubles drift. A pre-set grid crosses $\delta$ from +1 to $-8$ points with each of the other dials.

\textit{Checks.} Family-a worlds reproduce the real-data statistics: below-floor $\beta$ of 0.89 to 0.90 (real 0.92), in-support 0.86 (real 0.87), and a top-decile winner's curse of 1.41 in support (real 1.44). Family g, drawn from the drafter's own network class, does not (below-floor $\beta$ 0.78, top-decile curse 2.3). The worlds also miss two real-data statistics. Default worlds do not reproduce the real dip at rarest triple 1 to 4 (+0.47 against $-$1.77 observed, the worlds' 2nd percentile); under certify14 and stress the observed dip is consistent with the worlds. And when we draw outcomes from one world and refit the model, its 90\% intervals cover the truth 89 to 90\% of the time on human teams but only 63 to 89\% on search-chosen teams. The posterior understates its uncertainty where search goes, so the world results are less certain on search-chosen teams than their intervals suggest.

\subsection{What the Value Function's Errors Cost the Search}

\begin{table}[t]
\centering
\caption{Truth-leaf regret: true points gained when the search's value function is replaced by the world's truth, with the same search code and budget. For the MCTS policy, which plays without decision-time search, the counterpart is decision-time search with the true value, so its row also includes the value of search. Mean [95\% over worlds]. Opponents: MCTS, masked greedy, GD; 28 drafts per seat. Truth-leaf runs use the first 10 worlds of each family (5 for c), a subset of those in Table~\ref{tab:families}.}
\label{tab:regret}
\setlength{\tabcolsep}{2.5pt}
\begin{tabular}{lcccc}
\toprule
Agent & a, default & a, stress & c, default & g \\
\midrule
Search & +1.6 [0.2, 3.6] & +1.4 [0.1, 4.3] & +1.7 [0.4, 3.5] & 0.0 \\
Search, prior + floor & +1.3 [0.4, 3.0] & +1.3 [0.2, 3.0] & +1.4 [0.3, 2.3] & 0.0 \\
MCTS policy & +2.4 [0.9, 6.0] & +2.2 [0.2, 6.9] & +2.6 [1.1, 6.2] & +0.3 \\
\midrule
Worlds & 10 & 10 & 5 & 10 \\
\bottomrule
\end{tabular}
\end{table}

\paragraph{Truth-leaf regret} Give the search the true value of every team instead of its learned one, keep everything else fixed, and measure the gain in truth (Table~\ref{tab:regret}). Decision-time search loses 1.6 points to its value function's errors in the main family, and the MCTS policy played directly is 2.4 points below search with the true value. The spread is wide: in some worlds the errors cost almost nothing, in others up to 6 points. The loss is the same size under the stress dials, which make rare teams much worse than the model believes. When the world is drawn from the drafter's own network class (g), the loss vanishes, as it should.

\paragraph{No systematic overclaim} Truth-leaf regret counts every error, in either direction. To ask whether search is systematically fooled, we redraw every outcome inside a world, rebuild the statistics and features, refit the WP model, and search with it. Its claimed value then exceeds the true value of its teams by $-$0.6 points ($-$1.8 to +0.2). The search's errors cost it good moves, but they do not inflate what it believes it has won. The refitted pipeline also reproduces the two other main results: its value function is no worse a leaf than the original one (+0.15 points, $-$0.7 to +0.7), and a floor at play time costs it 2.7 points (1.5 to 4.0), the same as in the main rescoring.

\section{Staying on Support}
\label{sec:floor}

\begin{table*}[t]
\centering
\caption{Strength against support. True: paired difference from the MCTS policy in family-a worlds, true win-probability points, mean [95\% over 1,000 worlds]. Judged: mean win probability against the eight shared opponents under the consensus, as judged / after the support correction (SE $\le$ 0.15); the MCTS policy is itself one of the eight, so its row covers the other seven. Below: share of teams with a rarest triple under 25. All search rows use the role mask. Prior: the MCTS policy as trained, or trained with the floor inside its search. Floor at play: no pick may create a triple seen fewer than 25 times, unless no legal hero passes. Search: 200 simulations at decision time. Search rows pool seeds (MCTS policy five, floor-trained three); Section~\ref{sec:floor} and the supplement give eight separately trained seeds per policy.}
\label{tab:frontier}
\setlength{\tabcolsep}{5pt}
\begin{tabular}{llccccc}
\toprule
Drafter & Prior & Floor at play & Search & True vs.\ MCTS policy & Judged & Below \\
\midrule
MCTS policy & as trained & -- & -- & 0 & 58.1 / 58.3 & 63\% \\
\quad + floor & as trained & yes & -- & $-$2.6 [$-$4.1, $-$0.5] & 55.0 / 56.1 & 0.8\% \\
\quad + floor, search & as trained & yes & yes & $-$1.5 [$-$3.6, +1.3] & 55.3 / 56.6 & 1.1\% \\
MCTS policy & floored & -- & -- & +1.2 [$-$0.2, +2.6] & 56.1 / 56.8 & 45\% \\
\quad + floor & floored & yes & -- & 0.0 [$-$1.6, +1.5] & 55.1 / 56.6 & 0.3\% \\
\quad + floor, search & floored & yes & yes & \textbf{+0.6 [$-$1.0, +2.1]} & 55.2 / 56.8 & 0.3\% \\
\midrule
Highest win rate + floor & -- & yes & -- & -- & 53.2 / 54.7 & 9\% \\
Featureless MCTS + floor & as trained & yes & -- & -- & 55.1 / 56.1 & 0.3\% \\
\midrule
CQL (weight 2, enriched) & -- & -- & -- & $-$19.4 [$-$25.0, $-$10.9] & 42.6 / 43.8 & 14\% \\
IQL (expectile .7, temp.\ 10) & -- & -- & -- & $-$18.6 [$-$24.3, $-$10.3] & 42.5 / 43.3 & 23\% \\
MCQ (weight .5) & -- & -- & -- & -- & 46.1 / 45.0 & 92\% \\
Thresholded penalty (0.5) & -- & -- & -- & $-$12.8 [$-$16.4, $-$7.7] & 49.2 / 48.5 & 66\% \\
\bottomrule
\end{tabular}
\end{table*}

\begin{table*}[t]
\centering
\caption{The floor and the floor-trained prior across world families and dials. Each entry is the true difference from the MCTS policy as trained (masked, no floor), in win-probability points, mean [95\% over worlds]. Floor: the same policy with the floor at play time. Floor-trained + floor: the prior trained with the floor inside its search, with the floor at play time. + search: the same with 200 simulations of decision-time search. No mask: MCTS trained without the role mask. Families b (cross-team cells), d (map and tier cells), and e (patch drift) at default dials share family a's base draws and match its row to one decimal. Harshest dials: $k = 3$, $\delta = -6$, $a = 0.5$, $\sigma = 6$, doubled drift. Family e at three times the drift measured between real builds is shown.}
\label{tab:families}
\setlength{\tabcolsep}{5pt}
\begin{tabular}{lcrrrr}
\toprule
World family, dials & Worlds & Floor & Floor-trained + floor & + search & No mask \\
\midrule
a (main), default & 1,000 & $-$2.6 [$-$4.1, $-$0.5] & 0.0 [$-$1.6, +1.5] & +0.6 [$-$1.0, +2.1] & +1.7 [+0.2, +3.0] \\
a, certify14 & 1,000 & $-$2.3 [$-$3.8, $-$0.2] & +0.3 [$-$1.3, +1.9] & +1.0 [$-$0.6, +2.5] & +1.6 [+0.1, +2.9] \\
a, stress & 1,000 & $-$2.2 [$-$3.8, $-$0.1] & +0.4 [$-$1.2, +1.9] & +1.0 [$-$0.6, +2.5] & +1.6 [+0.1, +2.9] \\
c (four and five), certify14 & 1,000 & $-$1.8 [$-$3.4, +0.2] & +0.6 [$-$1.0, +2.2] & +1.3 [$-$0.2, +2.9] & +1.4 [$-$0.1, +2.7] \\
c, stress & 1,000 & $-$1.7 [$-$3.4, +0.4] & +0.7 [$-$1.0, +2.5] & +1.4 [$-$0.3, +3.2] & +1.4 [$-$0.1, +2.7] \\
c, harshest dials & 1,000 & $-$1.2 [$-$3.4, +1.3] & +1.0 [$-$1.1, +3.0] & +1.6 [$-$0.5, +3.5] & +1.6 [0.0, +3.0] \\
e, drift $\times$3 & 1,000 & $-$2.5 [$-$5.8, +1.2] & 0.0 [$-$6.9, +7.7] & +0.6 [$-$6.3, +7.8] & +1.6 [$-$1.2, +4.2] \\
g (drafter's network class) & 20 & $-$2.4 [$-$3.0, $-$2.0] & $-$1.1 [$-$1.6, $-$0.6] & $-$0.4 [$-$1.0, +0.05] & +1.3 [+0.9, +1.9] \\
\bottomrule
\end{tabular}
\end{table*}

\paragraph{A floor at play time costs real value} Even a search that knows the true value of every team loses 3.0 points (1.7 to 5.2) when it must obey the floor: the floor removes good teams, not only bad ones. On the policy as trained, the floor takes it from 63\% to 0.8\% of teams below the floor and costs 2.6 true points (Table~\ref{tab:frontier}). Across world families and dials the cost is 2.2 to 2.7 points in most settings (Table~\ref{tab:families}). It is smallest when four- and five-hero cells carry extra error, the settings built to make rare teams worst for the model: 1.7 to 1.8 points under certify14 and stress, and 1.2 ($-$1.3 to +3.4) under the harshest dials. There, and in family e at large drift, the interval crosses zero. The judges agree in direction: on the opponents common to both arms, the corrected consensus puts the floor's cost at 1.3 points.

\begin{figure*}[t]
\centering
\includegraphics[width=\textwidth]{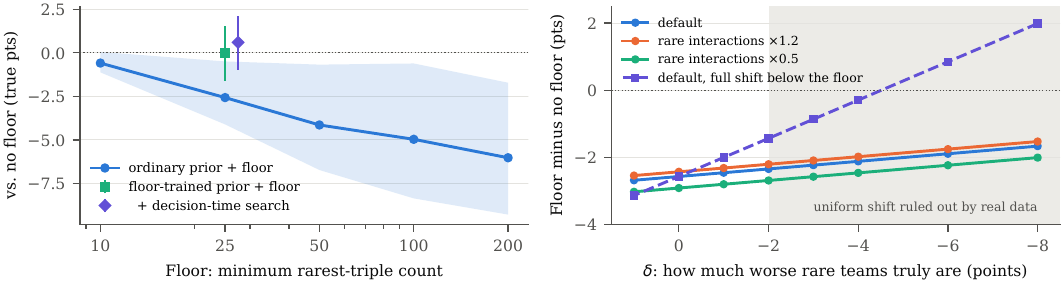}
\caption{What a play-time floor costs in truth (family a, pooled policies). Left: cost to the MCTS policy against the floor's threshold (mean and 95\% interval over 1,000 worlds), with the floor-trained prior under the floor at 25, alone and with decision-time search. Right: the cost of the floor at 25 as rare teams are made worse than the model believes ($\delta$, applied in full to teams far below the floor), in default worlds and with rare teams' interaction value scaled by 1.2 or 0.5 (means over worlds), and with the shift applied in full to every team below the floor (dashed). The shaded region marks uniform shifts larger than the real-data bound allows. Values are signed differences in true win-probability points; negative means the floor loses strength. The floor-trained markers are separate arms, compared with the same unfloored MCTS policy.}
\label{fig:floor}
\end{figure*}

\paragraph{A higher floor costs more} Fig.~\ref{fig:floor} (left) moves the threshold. At 10 the floor costs 0.6 points ($-$0.05 to 1.1) but still leaves 59\% of teams with a rarest triple under 25, so it barely binds. At 25 it costs 2.6 points, at 50 4.1, at 100 5.0, and at 200 6.0. The tighter the floor, the higher the cost, and no threshold we tried has a positive point estimate. Higher thresholds also bind less cleanly: when no legal hero passes, the fallback takes the best-supported hero, so the share of teams with a rarest triple under 25 rises from 0.8\% at a threshold of 25 to 6\%, 10\%, and 17\% at 50, 100, and 200. The floor also costs featureless MCTS 2.6 points and greedy value search 1.4. The win-rate drafter loses only 0.4 ($-$0.4 to 1.3) although 81\% of its teams are below the floor: the floor swaps one strong hero for another that is nearly as strong.

\paragraph{The floor pays only beyond what real data allow} On the pre-set grid the floor's cost falls as rare teams are made worse, from 2.6 points when the model is right to 1.7 when the dial makes rare teams 8 points worse than it believes (Fig.~\ref{fig:floor}, right). The dial is applied in full only far below the floor and shrinks toward the floor, so at the search's most common rare triple, seen 13 times, $\delta = -8$ amounts to about $-$1.5 points. On this grid the cost never reaches zero, and scaling rare teams' interaction value up or down shifts the curve without closing the gap. Section~\ref{sec:leaves} shows why: the search's rare teams sit just below the floor, where the dials, which grow with distance below the floor, barely act. A dial large enough to make those teams bad enough would also contradict the real-data measurements of Section~\ref{sec:wrong}, whose 95\% bound on extra overvaluation below the floor is about 2 points. To test this directly, we rerun the grid with the shift applied in full to every team below the floor, however close to it (Fig.~\ref{fig:floor}, right, dashed). The floor then pays once rare teams are about 4.5 points worse than the model believes, and not before: at a shift of $-$2 points, the most the real data allow for a uniform shift on human teams, it still costs the ordinary policy 1.5 points (1.0 to 2.0 over eight seeds). Every break-even point on the other planes lies between $-$3.5 and $-$7 points.

\paragraph{Training the prior under the floor reduces its cost} A policy trained without the floor learns to want what the floor forbids, and under the floor it falls back to second choices it never learned to make well. We therefore train the MCTS policy with the floor inside its search, at the root, at expansion, and in our own rollout picks, with the same configuration and budget as the ordinary policy, so that it learns to draft well within support. To separate the effect from training noise we trained eight seeds of each policy and scored every seed in the same 1,000 worlds (intervals in this paragraph and the next are 95\% over seeds; supplement). The floor costs the floor-trained policy 1.1 points (0.9 to 1.3) and the ordinary policy 2.6 (2.1 to 3.2), and every floor-trained seed pays less than every ordinary one (0.8 to 1.5 against 1.6 to 3.5).

\paragraph{The floor-trained prior is also stronger} Without the floor, the floor-trained policy beats the ordinary one by 1.6 points (0.4 to 2.8), with seven of eight seeds above the ordinary mean, although it still puts 45\% of its teams below the floor: training under the floor changes what it prefers without confining it. With the floor on, it is +0.5 points ($-$0.8 to 1.8) against the ordinary policy without the floor. Adding 200 simulations of decision-time search gives +0.8 ($-$0.5 to 2.0), with 0.3\% of teams below the floor. That is no detected cost rather than proven equivalence; the interval allows a loss of half a point. If rare teams are worse than the model believes, the floor-trained arms gain. With every below-floor team 2 points worse, the floor costs the floor-trained policy 0.3 points (0.2 to 0.5), and the on-support search beats the unconstrained ordinary policy by 2.0 (0.7 to 3.3).

The pooled policies of Table~\ref{tab:frontier} give the same picture (0.0 and +0.6 points over worlds). The floor-trained arms hold their place in every family built from the outcome model, including patch drift, where the intervals widen because the world moves after training (Table~\ref{tab:families}). These families share their base draws, so at default dials they give the same answer; they differ only where their extra cells are turned up. The result is not universal: in family g, drawn from the drafter's own network class, the on-support search loses 0.4 points ($-$0.05 to 1.0) and the floor-trained policy with the floor loses 1.1 (0.6 to 1.6).

\begin{figure*}[t]
\centering
\includegraphics[width=\textwidth]{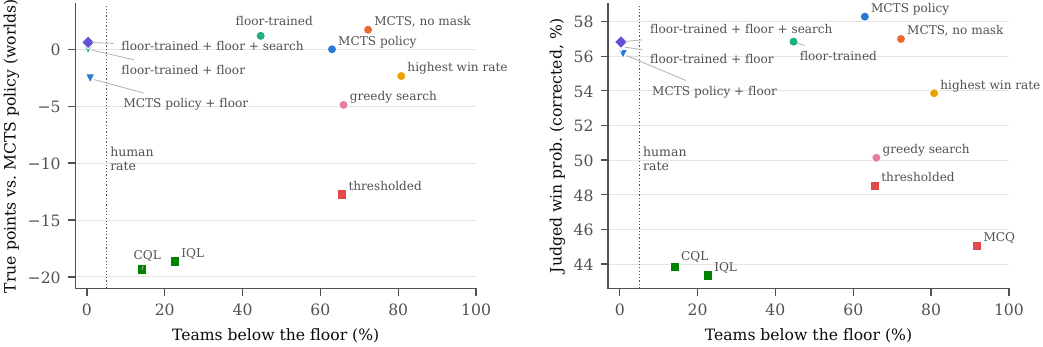}
\caption{Strength against support for the drafters of Table~\ref{tab:frontier}, plus unmasked MCTS, greedy value search, and the win-rate drafter. Left: true strength in family-a worlds, relative to the MCTS policy. Right: judged win probability against the eight shared opponents, after the support correction (the MCTS policy, itself one of the eight, against the other seven). MCQ has no true-strength result and is omitted on the left. The dotted line marks the human rate of teams below the floor (5\%).}
\label{fig:frontier}
\end{figure*}

Fig.~\ref{fig:frontier} puts the drafters on one plane. In truth (left), only the floor-trained arms reach the upper left, strong and on support. The plain floor moves the MCTS policy left and down, and the offline RL methods sit far below both. The judges (right) agree with the worlds except at the top, where they rank the unconstrained policy first and the on-support search below it; in the worlds the on-support search, the floor-trained policy, and MCTS without the mask are all ahead of it. The judges are trained on human teams and reward the rare teams the floor removes (Section~\ref{sec:refs}). We read the truth column for the floor's cost and report both.

\paragraph{An uncertainty penalty does not bring it back} The standard search-time remedy is to score leaves pessimistically. We replace the leaf value of unmasked decision-time search (the MCTS policy without the role mask, with 200 simulations) with the mean of a 20-member out-of-fold ensemble, minus one or two standard deviations. The ensemble members differ in architecture, data half, seed, and feature groups (supplement). The search barely moves: 71 to 72\% of its teams stay below the floor, against 71\% with the plain value, and its judged score is unchanged (56.5 against 56.6, corrected). In the worlds the penalty costs 0.2 points ($-$0.05 to 0.5). It neither keeps the search on support nor makes it stronger.

\paragraph{Pessimism does not solve it} The offline RL methods we tested split into two groups (Table~\ref{tab:frontier}; Monte Carlo and bootstrapped variants and their sweeps are in the supplement). CQL and IQL stay near human play. They keep most of their teams exactly as humans fielded them, are judged at 42 to 44\%, and lose about 19 true points to search in the worlds (11 to 25). They still put 14 to 23\% of their teams below the floor. MCQ and a thresholded CQL penalty, which do not pull toward the logged action, leave human play: MCQ puts 92\% of its teams below the floor and is judged at 45\%, and the thresholded penalty puts 66\% there and is judged at 48.5\%. No setting we tested reaches the strength of search, and none stays on support as well as floored search. A behavior penalty cannot tell a team that is rare because it is bad from one that is rare because few people have tried it, and on real games rare human teams are no more overrated than common ones.

\section{Discussion}
\label{sec:discussion}

\textit{Why the floor does not pay here.} A floor pays when the teams it removes are worse than the value function believes by more than the value it gives up. In the worlds we tested, the floor costs strength without a matching reduction in the value function's error. The search's rare teams are near the edge of the data, built from triples humans have played, and a value function built from pair and composition statistics can interpolate to them. And the real-data error that remains is a winner's curse that is the same size on common teams, so removing rare teams does not remove it. A floor is more likely to pay where search reaches far outside the data, as MCQ does with 28\% of its teams built on a triple never seen, or where the value function extrapolates with no structure to lean on.

\textit{What the measurements are worth elsewhere.} The three measurements need only logged decisions, held-out outcomes, and a fitted outcome model, so they apply wherever a learned evaluator is searched: games, molecule and protein design, and language models optimized against a learned reward. The first is a support distribution for the optimizer's own choices against the data's. The second is a held-out test of the evaluator's error by support, with a backtest that trains on older data and scores on newer. The third is a world fit to the data in which the truth is known. Together they separate two questions that pessimism methods fold into one: whether the evaluator is more wrong off support, and what its errors cost the optimizer wherever they are. Here the evaluator was not detectably more wrong off support, and its errors still cost the search about 1.6 points, through which teams it chose rather than through a systematic overclaim. A support constraint addresses only the first kind of error.

\textit{Training under the constraint.} On the ordinary policy the floor costs 2.6 points; on the floor-trained policy it costs 1.1, and less in every one of eight training seeds. Training with the floor inside the search teaches the policy which supported teams are good, so the constraint at play time costs less, and the prior is stronger even without it. The same idea applies wherever a constraint is added at deployment to a policy trained without it.

\section{Limitations and Future Work}

\textit{Real-data tests use human-chosen teams.} Every real-game test in Section~\ref{sec:wrong} uses teams that human players chose. Search-chosen teams at the same rarity are less well determined by the data, and the 95\% upper bound on extra off-support error, about 2 points, is close to what the floor costs. That claim rests on the simulated worlds. Outcomes of games played with recommended teams test it directly; a prospective test of this kind is under way.

\textit{Worlds encode a model class.} The worlds are drawn from an explicit model fit to the same games, and in family g the floor does cost the on-support drafter about half a point. They can represent any off-support error the dials describe, but not every possible one. For this reason we report the stress dials and the family drawn from the drafter's own network class next to the default. In default worlds the real dip at rarest triple 1 to 4 sits at the 2nd percentile of the worlds' range.

\textit{Judges are learned.} The judges share no data with any agent, but they are value functions trained on human games, with their own support bias. Their raw and corrected scores differ from the truth column of Table~\ref{tab:frontier} in level and, for the floor-trained arms, in order.

\textit{Players.} The value function, the search, and the judges are all hero-level. A ranked match is played by ten particular people, and who plays each hero may also matter. A drafter that knows its players faces a second axis of support, the player on each hero, and its own version of the questions in this paper. We leave that to future work.

\section{Conclusion}

Search over a learned value function does leave its data, and in MOBA drafting it does so in most games. On real games, a carefully built value function is no more wrong about rare human teams than about common ones. Its main measured weakness is a winner's curse on the human teams it likes most, which a support floor does not target. In ground-truth worlds its errors cost the search about a point and a half, and a plain play-time floor costs more than it saves. Training the prior under the same floor cuts that cost from 2.6 points to 1.1, less in every one of eight training seeds, and with prior and floor together the drafter stays on support with no detected net cost in the main world family.

Three lessons carry over to other settings where a learned evaluator is searched. Measure the value function's error by support directly, on held-out data and with a backtest, before paying for pessimism. Separate the error that depends on support from the winner's curse that does not, since they need different remedies. And when a support constraint is needed, train under it rather than only imposing it at play time.

\section*{Ethics}

The expert study received an exempt determination from the Columbia University Institutional Review Board (protocol IRB-ACYY3193, exempt category 2).

Raters were recruited among experienced players and gave electronic consent before rating, after a notice that stated that some drafts were machine-generated and how their data would be handled. Each rater who completed all 240 items received a flat US\$100 prepaid card, whether or not they passed the analysis gates; partial sets were not paid, as stated at recruitment. Raters appear in the analysis only under code names. All game data are public match records; players are identified only by numeric account identifiers and were not contacted.

\section*{Acknowledgment}

We thank Heroes Profile (heroesprofile.com), whose API supplied our replay corpus. Heroes of the Storm is a trademark of Blizzard Entertainment.

Generative AI (Claude Code, Anthropic; Codex, OpenAI) has been used in the preparation of portions of this document, both in manuscript editing as well as in code writing, design, and data analysis. The authors directed the research, chose the experiments and framing, checked the results and text, and take full responsibility for the content.

\section*{Conflicts of Interest}

The authors declare no conflicts of interest.

\end{document}


\title{Supplementary Material for\\ Searching Beyond the Data: Value-Function Exploitation\\ and the Cost of Staying on Support in MOBA Drafting}
\author{Max~Segan, Ernest~Pysher, James~Frankel, and Hod~Lipson%
\thanks{M. Segan is with the Department of Computer Science and H. Lipson with the Department of Mechanical Engineering, Columbia University (e-mail: m.segan@columbia.edu; hod.lipson@columbia.edu). E. Pysher and J. Frankel are independent researchers (e-mail: ernestpysheriv@gmail.com; james@frankel.tv).}}

\maketitle

\section{Support Measures and the Judges' Support Correction}
\label{sec:supportsupp}

\paragraph{Measures} Every support statistic is computed against the 3.73 million training-split teams (both sides of 1.87M games), ignoring map, tier, and opponent. Rarest pair, triple, and four: the smallest count among the team's 10 pairs, 10 triples, or 5 four-hero subsets. Exact five: the count of the team itself. Pair affinity: the mean pointwise mutual information of the team's 10 pairs, $\log[P(a,b)/P(a)P(b)]$ over team co-occurrence. Role composition: the count of the team's sorted role multiset. Matchup: the smallest count among the 25 cross-team hero pairs of the game, in games. Each is put on one scale, the share of held-out human teams strictly rarer, and the floor is the 5th percentile. Human 5th and 50th percentiles: rarest triple 23 and 215, rarest pair 1,049 and 4,669, role composition 11,950 and 413,074, matchup 1,369 and 4,102, pair affinity $-$0.26 and $-$0.09. The exact-five and rarest-four measures have many ties at zero, so 36.6\% and 7.8\% of human teams fall below their floors.

\paragraph{Judges' support bias} The judges are trained on human games, so we audit them on real teams by support. On 1M training-split games scored with leave-one-out support and 76,969 validation and test games, the consensus underpredicts how badly the rarest 1\% of teams by rarest triple do by 4.0 points (gN 3.0, RN 4.1), the 1--2.5\% bin by 1.6, and overpredicts the top half by 0.2. The gap does not come mainly from who fields such teams: on 266K games where all ten players have a Heroes Profile MMR recorded before the game, controlling for the MMR difference, MMR by pick position, and party moves the bottom-5\% pair-affinity effect from $-$1.01 to between $-$0.96 and $-$0.86 points. Holding a four-hero core fixed and varying only the fifth hero, a 1 SD rise in pair affinity is worth +0.56 points in reality and +0.16 under the consensus, so the bias survives within a core.

\paragraph{Correction} We correct each judge with an offset binned by rarest triple, rarest pair, and pair affinity, fit by 5-fold cross-validation on the 1M training games (log loss 0.68330 uncorrected, 0.68306 binned). It removes the tail gap in cross-validation (rarest-1\% triple $-$3.8 $\to$ $-$0.2 points; rarest-1\% affinity $-$3.0 $\to$ +0.5). It over-corrects the affinity tail by about half a point, and it is partly aligned with the floor by construction, so the main paper reports judged scores both before and after it and reads the floor's cost from the ground-truth worlds.

\section{Ground-Truth Worlds}
\label{sec:worldssupp}

\paragraph{World model} The base model is
\[
\mathrm{logit}\,P(\text{team 0 wins}) = s(T_0) - s(T_1) + \textstyle\sum_{a \in T_0, b \in T_1} C_{ab},
\]
with $s(T) = \sum_{h \in T}(w_h + V_{m,h} + U_{t,h}) + \sum_{a<b \in T} S_{ab} + \sum_k \sum_{\{a,b,c\} \subset T} E_{ak}E_{bk}E_{ck}$ (hero, hero-map, hero-tier, pair synergy, and a rank-8 three-hero term) and $C$ an antisymmetric matrix of cross-team counters. Priors are Gaussian; the pair and embedding prior scales (0.05 and 0.4) and the rank were chosen on the validation split. The posterior is a Laplace approximation at the MAP with the Gauss-Newton precision. Held-out log loss: validation 0.67729, test 0.67326, post-snapshot 0.67731, against 0.6782 for the WP model on the post-snapshot games.

\paragraph{Higher-order layers} Independent cell effects are added for every triple, quartet and five of a team, every (pair, opposing hero) cell, and every hero-map-tier cell. Each layer's scale is chosen by type-II maximum likelihood; all come out small (at most 0.005 logit), so the data rule out large residual effects on average. Seen cells take their posterior; unseen cells draw from the prior. Each world draws its cells from a counter-based hash of the cell, the world, and the layer, so a world is a fixed function of any team.

\paragraph{Dials} A team's grade $g(m) = \mathrm{clip}(1 - \log(1+m)/\log(1+F), 0, 1)$ is 1 for an unseen cell or team and 0 at or above the floor $F$. Dials act in proportion to the grade: $k$ scales layer cell SDs by $1 + (k-1)g$, $\delta$ shifts team strength by $\delta g$ points, $a$ scales the family's interaction class by $1 + (a-1)g$, and $\sigma$ adds team-level noise of SD $\sigma g$ points. Defaults: $k = 1$, $\delta = 0$, $a = 1$, $\sigma = 0$. Named settings: qm ($\delta = -1$), certify14 ($\delta = -2.7$), atten ($a = 0.84$), sigma3, stress ($k = 2$, $\delta = -3$, $\sigma = 3$, drift doubled), and stress\_hard ($k = 3$, $\delta = -6$, $a = 0.5$, $\sigma = 6$). The break-even grid crosses $\delta \in \{+1, 0, \ldots, -8\}$ with $\sigma \in \{0, 3, 6\}$, $k \in \{1, 1.5, 2, 3\}$, and $a \in \{1.2, 1, 0.85, 0.7, 0.5\}$. All were frozen before any agent was scored. Base draws are shared across families for a given world, so families are paired.

\paragraph{Checks} On 200 certify14 worlds, real statistics fall inside the worlds' range: below-floor $\beta$ at the 64th percentile, in-support $\beta$ at the 72nd, the top-decile winner's curse at the 36th (below) and 56th (in support). The dip at rarest triple 1--4 sits at the 23rd percentile under certify14 and the 2nd under defaults. Recovery: drawing outcomes from worlds 900--902 and refitting, 90\% intervals cover the truth 89--90\% of the time on human teams and 63--89\% on search-chosen teams (z-score SD 1.06--1.49).

\paragraph{Break-even} The floor's paired cost against the MCTS policy in family a, by full-grade shift $\delta$: $-$2.68 at $\delta = +1$, $-$2.57 at 0, $-$2.34 at $-$2, $-$2.12 at $-$4, $-$1.89 at $-$6, $-$1.67 at $-$8. The $\sigma$ and $k$ planes are within 0.01 of these; $a = 0.5$ makes the floor cost 0.3 more and $a = 1.2$ 0.1 less. No cell on any plane reaches break-even under this graded rule. With the shift applied in full to every team below the floor (the flat grade), the floor's cost is $-$2.57 at $\delta = 0$, $-$1.43 at $-$2, $-$0.29 at $-$4, and +0.85 at $-$6, with break-even at $\delta = -4.5$; the $\sigma$ and $k$ planes break even between $-$4.4 and $-$4.5, and the $a$ plane between $-$3.5 ($a = 1.2$) and $-$7.1 ($a = 0.5$).

\section{Exploration and the Value of Information}
\label{sec:explore}

A deployed drafter could recommend uncertain teams on purpose, so that the games its users play teach the value function about them. We simulate this inside the worlds: 4,000 rounds, four organic games per round, a share $c$ of players who follow the drafter, and a learner from the WP model's class refit as games arrive. Policies: always exploit; break near-ties toward uncertain teams; anneal the floor; a lower confidence bound; $\epsilon$-greedy; top-two Thompson sampling; and refitting without deliberate exploration. At $c = 20\%$, every policy is within $-$0.034 to +0.011 true points per round of pure exploitation. Recommended teams are followed rarely (0.5 to 0.7\% of games), and one pair needs 1.8 weeks to reach 25 games at 2,000 followed drafts per week, but 147 weeks at 25. A year of 2,000 drafts per week would gain at most 0.13 to 0.22 points per draft. Deliberate exploration does not pay at realistic traffic; collecting outcomes on recommended teams is still the direct test of the main paper's off-support claims.

\section{Fused CUDA Kernels for MCTS Training}
\label{sec:cuda}

The fused CUDA implementation described here ran every MCTS training run in the main paper, up to 800 simulations per move and 1M episodes. AlphaZero-style MCTS training is computationally demanding: each episode requires $\sim$8 MCTS searches of 200 simulations each, with a neural network forward pass at every leaf node. Standard implementations incur Python interpreter overhead, CUDA kernel launch latency, and host-device synchronization for each of the $\sim$3,200 forward passes per episode.

We developed a fused CUDA implementation that runs entire MCTS episodes inside a single kernel launch. One thread block (256 threads) executes one complete draft episode: thread~0 drives all sequential logic (UCB selection, action sampling, backpropagation) while all 256 threads cooperate on matrix multiplications during neural network forward passes via shared memory. The forward pass functions (\texttt{linear\_layer}, \texttt{batchnorm\_relu}, \texttt{residual\_block}) are compiled as \texttt{\_\_device\_\_} functions callable from within the kernel with zero launch overhead. Two additional optimizations compound with the fused kernel:
\begin{itemize}
    \item \textbf{Split backbone from heads}: Value symmetrization runs only the small value head (257$\to$128$\to$64$\to$1) twice rather than the full backbone, halving symmetrization cost.
    \item \textbf{Pipelined generation}: Episode results are written into pre-allocated ring buffers via C++ \texttt{memcpy}; a generation thread launches GPU batches asynchronously while the main thread performs gradient updates.
\end{itemize}

\paragraph{Search with opponent expansion.} A shallow search expands a node only at our team's turns, so its tree covers the current own-pick block. Every MCTS agent in the paper trains with a search that also expands the opponent's turns: an opponent node is a chance node whose GD reply distribution is computed once and cached, its children are added with progressive widening (at most $\lceil\sqrt{N}\rceil$ after $N$ visits), and each simulation creates at most one new own decision, which costs a policy forward pass. Trees are allocated per episode ($3 \times$ simulations $+ 8$ nodes). We measured the kernel with the protocol of the controlled comparison below (128 concurrent episodes per launch, one warm-up launch, then 768 timed episodes; the paper's operating-point agent and leaf WP) on one NVIDIA RTX 5090. With this search the kernel runs 63 episodes/second at 200 simulations per move and 33 at 400; with the shallow search on the same GPU, 216 and 112. On the RTX 5090, the deeper search reduces throughput by a factor of 3.4.

\paragraph{Controlled baseline (shallow search).} For a controlled speedup claim we built a tuned host-tree baseline in the style of KataGo's batched-evaluation pipelines: 64--256 CPU threads each running one episode's tree, leaf evaluations batched across threads onto the GPU via cuBLAS GEMM over multiple streams, reproducing the production kernel's exact per-simulation work (policy priors at expansion, Generic Draft rollout to terminal at every leaf, symmetrized enriched-WP scoring). At equal work \textit{and} equal search semantics (leaf batch 1; first-turn root-visit total-variation distance from the kernel 0.047, at the 0.044 kernel-vs-kernel seed-noise floor; win rate vs.\ GD 0.761 vs.\ the kernel's 0.757), the host pipeline sustains 68 ep/s against the fused kernel's 151 ep/s on the same GPU: the fused kernel is 2.2$\times$ faster, because per-simulation host-device round trips cannot be amortized without within-episode leaf batching. Virtual-loss leaf batching does push the host pipeline to 163 ep/s (1.07$\times$ the kernel), but only by degrading the search itself: at leaf batch 32 the win rate vs.\ GD falls from 0.757 to 0.531 and the visit-distribution TV rises to 0.62; softening virtual loss to 0.25 recovers only 0.712. Per-component profiling shows the GD rollout evaluation dominates ($\sim$75\% of the host pipeline's evaluator wait time; policy priors and WP scoring share the rest). This comparison covers only the shallow search, the only one the host pipeline implements, and was measured on RTX PRO 6000 GPUs with their host CPUs. It was not repeated for the full search: the host pipeline runs 256 CPU threads, and on the RTX 5090 machine's 16 cores it reaches 9 episodes/second, so its speed there says more about that machine's CPU than about the kernel. The fused kernel also offers 0.12~s startup, a deterministic $\sim$270~MB footprint, and no host thread pool or Python runtime at inference.

\section{Validity of the Interaction Metrics}
\label{sec:metricvalid}

\begin{table}[t]
\centering
\caption{Strength-adjusted validity of the interaction metrics: actual win rate in the top minus the bottom metric decile (pp), and the number of increasing adjacent decile steps (of 9). Deciles are computed within strata of skill tier $\times$ own-team $\times$ opposing-team mean-hero-WR quintiles and count-weighted across strata (95\% CIs about $\pm$1.1pp on the test set and $\pm$0.4pp post-snapshot). Counts are team-perspective rows, with two rows per replay. ``Self-incl.'': the statistics contain the scored games' own outcomes.}
\label{tab:metricvalid}
\setlength{\tabcolsep}{2.5pt}
\small
\begin{tabular}{llccc}
\toprule
Statistics & Rows & Synergy & Counter & Self-incl. \\
\midrule
External & test (77,738) & +11.3, 9/9 & +11.0, 8/9 & yes \\
External & post (521,078) & +8.8, 9/9 & +9.2, 9/9 & no \\
Own train split & test & +9.5, 7/9 & +14.9, 7/9 & no \\
Own train split & post & +7.5, 9/9 & +12.6, 9/9 & no \\
\bottomrule
\end{tabular}
\end{table}

The main paper's interaction-awareness metrics are built from the pairwise synergy and counter deltas (Section~\ref{sec:bench}), which subtract from each observed pair win rate an independence expectation formed from the two heroes' individual win rates. We validate that these deltas carry real outcome signal by scoring each side of real drafts with the exact metric implementations against the game's actual outcome. Pooled naively, the decile curves are not monotone: subtracting the individual-strength expectation makes the deltas strongly anti-correlated with team hero strength (team synergy delta against team mean hero win rate), so teams of individually strong heroes receive negative deltas while still winning more. We therefore control for that component, computing deciles within tier and own/opposing team-strength strata.

On the test replays the external statistics contain each replay's own outcome (first row of Table~\ref{tab:metricvalid}). On 260,539 post-snapshot games, which were uploaded after the statistics were pulled and so cannot be in them, the same statistics give spreads of +8.8pp (synergy) and +9.2pp (counter), both monotone across all nine steps. The self-including spreads are 2.5pp and 1.8pp larger, about 30\% and 20\% of the clean values, though the two figures come from different game sets. Statistics built only from our training split give the same qualitative result on both game sets. The interaction deltas measure an outcome-relevant property of a composition beyond individual hero strength, which is the component they are designed to isolate. 

\section{Statistics Behind the Interaction Metrics}
\label{sec:statsens}

The interaction metrics read pairwise win-rate statistics, and the main paper reports them with the community aggregates. Does the ordering of agents depend on that choice? We score the same benchmark drafts of every MCTS configuration (Table~\ref{tab:mcts}; root temperature 1) with two statistics sets: our own training-split statistics, which the out-of-fold value functions also read, and the external aggregates. Same policies, same drafts; only the lookup tables differ.

\begin{table}[t]
\centering
\caption{Synergy and counter responsiveness of the MCTS configurations' benchmark drafts under our training-split statistics (Own) and the external aggregates (Ext.). Mean over seeds.}
\label{tab:statsens}
\setlength{\tabcolsep}{3.5pt}
\small
\begin{tabular}{lcccc}
\toprule
& \multicolumn{2}{c}{Synergy} & \multicolumn{2}{c}{Counter} \\
Config & Own & Ext. & Own & Ext. \\
\midrule
B & +0.93 & +1.06 & +.074 & $-$.028 \\
F & +1.08 & +1.29 & +.087 & $-$.020 \\
J & +1.06 & +1.31 & +.120 & +.023 \\
E (1M ep.) & +1.20 & +1.21 & +.082 & $-$.022 \\
K (no feat.) & +0.88 & +1.24 & +.116 & +.016 \\
B, in-sample & +1.14 & +1.07 & +.086 & $-$.015 \\
F, in-sample & +1.74 & +1.56 & +.091 & +.006 \\
J, in-sample & +2.38 & +2.02 & +.094 & +.009 \\
\bottomrule
\end{tabular}
\end{table}

The counter ordering of the eight configurations is nearly the same under both statistics sets (Spearman 0.98), with a level shift of about 0.1. The synergy ordering is not (Spearman 0.55). Most of the disagreement is the featureless agent K, which is the least synergistic configuration under our statistics and mid-table under the external ones, and the in-sample agents, whose value function read our statistics and which score higher under them. Synergy levels therefore carry information about which statistics an agent was built on; we compare synergy across agents only under one statistics set, and the main paper uses the external aggregates for every row.

\section{External Validation on Coordinated-League Drafts}
\label{sec:ngs}

The main paper's behavior policy is uncoordinated ranked-ladder play, and every evaluation in it also uses uncoordinated play (ranked or Quick Match games). As an external test, we evaluate the frozen snapshot-trained win-probability models on drafts from the Nexus Gaming Series (NGS), an amateur organized league in which persistent five-player teams draft together with dedicated captains. Scanning the Heroes Profile NGS archive yielded 11{,}132 games, of which \textbf{11{,}034} pass filtering (patch 2.55, complete 16-step draft, known winner), spanning 2021--2026 across ten season clusters, the same patch era as our ladder corpus.

\paragraph{Coordinated drafts are \textit{more} meta-concentrated.} A natural hypothesis is that coordinated teams, freed from solo-queue comfort picking, explore more of the hero pool. The data show the opposite, at both the hero and the composition level. Hero level: NGS pick entropy is 5.92 bits vs.\ the ladder's 6.25 (effective pool 60.5 vs.\ 75.9 heroes), the top-10 heroes take 33.5\% of picks vs.\ 25.1\%, and the off-meta share (picks outside the ladder top-30) is 39.8\% vs.\ 43.3\%; hero pick-rate rank correlation between the populations is 0.74. Composition level: NGS fields 78 distinct role compositions vs.\ 238 on ladder, with composition entropy 2.69 vs.\ 3.06 bits, the top-5 compositions covering 84.3\% vs.\ 78.6\% of teams, and 90\% coverage reached in 9 vs.\ 12 compositions. The modal composition is the same standard core in both populations (bruiser/healer/double-ranged/tank: 53\% vs.\ 50\%); NGS's second-most-common composition is a double-tank variant (12.3\%). Coordinated teams appear to converge on a known-good playbook, while ladder ``diversity'' partly reflects uncoordinated comfort picking and error. This fits the main paper's description of the behavior policy as shaped largely by social habit. The counts of distinct heroes and compositions depend on sample size (11K vs.\ 1.94M games); the entropies and top-$k$ shares carry the comparison.

\paragraph{Win-probability transfer.} Scored at the game level with symmetrized predictions, the out-of-fold enriched model reaches 55.4\% accuracy on NGS (95\% CI $\pm$0.9pp) against 57.5\% on the held-out ladder test set; the out-of-fold naive model also scores 55.4\%. NGS games are in neither our corpus nor the Storm League aggregates, so no model has self-inclusion here. The ladder-to-NGS drop is 2.1--2.2pp for both models. Calibration slope on NGS is 0.85 for the enriched model and 0.91 for the naive model. Splitting NGS drafts into quartiles by off-meta pick share shows no trend: the enriched model scores 56.8\%, 54.3\%, 55.7\%, and 54.7\% from most- to least-meta quartile (about 2,760 games each, CI $\pm$1.9pp), and the naive model 57.0\%, 53.9\%, 55.9\%, and 54.8\%. 99.98\% of NGS team compositions fall within the composition sets observed in the ladder-era aggregate statistics, so this shift re-weights familiar structures and does not reach the never-observed compositions. Persistent team-strength differences between the organized teams, which the models never observe, also limit how directly the ladder and league accuracies compare.

\section{Draft Concentration and Out-of-Distribution Extrapolation}
\label{sec:concentration}

How far do the agents drift from human play? We measure the operating-point agent (F, 5 seeds $\times$ 1,000 benchmark drafts, mid tier, 200 inference simulations, root temperature 1) and, for comparison, J (800 training simulations, 5 seeds).

\paragraph{Concentration} Each F seed uses 22--26 distinct heroes at 3.24--3.52 bits of pick entropy; pooling the five seeds gives 33 heroes and 3.54 bits, so the seeds converge on one head. Its seven favorite heroes are Samuro, Rehgar, Hogger, Deathwing, Falstad, Artanis, and Abathur; J's head overlaps heavily (Hogger, Samuro, Rehgar, Deathwing, Illidan, Cho, Auriel) and is slightly more diverse (26--31 heroes per seed, 3.61 bits pooled).

\paragraph{Are the favorites' interactions real?} For all 21 pairs and 35 trios of the seven favorites we compare the win rate of real teams containing them with the independence expectation from the heroes' own win rates (the paper's synergy convention). In our training split the favorite pairs beat independence by +0.36pp ($n$ = 273,540 team-games, $z$ = 3.8) and the trios by +1.16pp ($n$ = 17,771, $z$ = 3.1). On 414,926 post-snapshot games (including the drifted 2.55.17 builds), which no model saw, the pairs beat it by +0.09pp ($n$ = 60,675, $z$ = 0.4) and the trios by +0.75pp ($n$ = 4,013, $z$ = 0.9). Neither edge is significant forward. J's favorites do somewhat better forward (+0.55pp pairs, $z$ = 1.9; +1.90pp trios, $z$ = 1.1), still short of significance.

\paragraph{Where the agent leaves the data} None of F's ten most frequent five-hero teams (together 41.2\% of its drafts) occurs even once in the 1.94M-game corpus. One level up it stays in familiar territory: 96.9\% of its drafts land on role compositions admitted to our own composition table (at least 50 games in the tier label) and 47.4\% on compositions with at least 1,000 games.

\section{Ensemble Epistemic Uncertainty of the Value Model}
\label{sec:ensemble}

Do functionally different value models agree on the drafts the tournament judged? We train a 20-member out-of-fold ensemble on our out-of-fold features (validation early stopping): 8 full enriched models varying architecture (256$\to$128 or 512$\to$256$\to$128), training-data half, and seed; 9 models each dropping one enriched feature group; and 3 identity-only models. For each draft we compute the variance of the 20 symmetrized win probabilities.

\begin{table}[t]
\centering
\caption{Per-draft variance of the 20-member out-of-fold ensemble. Ratio: median relative to held-out human drafts; $>$p95: share of drafts above the human 95th percentile.}
\label{tab:ensvar}
\small
\setlength{\tabcolsep}{4pt}
\begin{tabular}{lrrr}
\toprule
Draft set & Med.\,$\times10^{3}$ & Ratio & $>$p95 \\
\midrule
Held-out human ($n$=2000) & 0.41 & 1.00 & (5.0\%) \\
F agent ($n$=2000) & 0.56 & 1.39 & 12.1\% \\
Synthetic degenerate ($n$=500) & 1.53 & 3.76 & 60.2\% \\
Degenerate probes ($n$=6) & 2.29 & 5.65 & 6/6 \\
\midrule
\multicolumn{4}{l}{\textit{Tournament drafts per strategy ($n$=2000 each)}}\\
Gourdeau estimator & 0.87 & 2.16 & 33.8\% \\
MCTS, no features & 0.83 & 2.04 & 31.5\% \\
MCTS & 0.77 & 1.89 & 26.5\% \\
Enriched greedy & 0.76 & 1.87 & 27.2\% \\
Constrained MCTS & 0.75 & 1.84 & 23.7\% \\
Constrained greedy & 0.74 & 1.82 & 26.1\% \\
TCQL & 0.69 & 1.70 & 25.6\% \\
CQL enriched & 0.55 & 1.37 & 12.3\% \\
CQL naive & 0.53 & 1.31 & 10.8\% \\
GD & 0.53 & 1.31 & 12.2\% \\
G\&A discriminator & 0.52 & 1.29 & 11.8\% \\
\bottomrule
\end{tabular}
\end{table}

The operating-point agent's benchmark drafts draw near-human disagreement (1.39$\times$ the human median, 12.1\% beyond the human 95th percentile). Degenerate drafts are sharply elevated (3.8$\times$ for synthetic unseen compositions, 5.7$\times$ for the fixed probes). The eleven learned strategies' tournament drafts sit between 1.29$\times$ and 2.16$\times$, below the degenerate regime. The anchored strategies are priced most confidently and the strategies that search a value function off the human manifold highest, with TCQL between. The tournament's verdicts were therefore issued where the model family mostly agrees, with the least agreement on the Gourdeau estimator's drafts (33.8\% beyond the human 95th percentile). Seeds that share one set of statistics share its errors, so their agreement does not validate them, as Section~\ref{sec:split} shows.

\section{Leakage Mechanism: A Within-Snapshot Split Experiment}
\label{sec:split}

To isolate how in-sample statistics affect a value function and the search that optimizes it, without any balance drift, we split the ranked snapshot by a salted replay-level hash into halves A and B. Proxies, their statistics, their early stopping, the behavioral-cloning prior and opponent of the search, and any search against them use only half A; references are built only from half B (three out-of-fold enriched judges, a hero-identity judge, and a cross-fitted realized-outcome index), plus the post-snapshot index and the Quick Match judge.

\begin{table}[t]
\centering
\caption{Held-out quality on half B of value functions trained on half A. ``Own-split val.'': what a pipeline with in-sample statistics reports about itself (last epoch).}
\label{tab:split}
\setlength{\tabcolsep}{3.5pt}
\begin{tabular}{lcccc}
\toprule
Proxy statistics & Games & Own-split val. & Held-out & Slope \\
\midrule
in-sample, 1/8 of A & 119K & 67.8\% & 53.3\% & 0.17 \\
in-sample, all of A & 952K & 59.7\% & 56.2\% & 0.62 \\
out-of-fold, all of A & 952K & 56.2\% & 57.0\% & 1.02 \\
hero identity only & 952K & 56.5\% & 57.1\% & 1.04 \\
\bottomrule
\end{tabular}
\end{table}

Table~\ref{tab:split} shows the mechanism. In-sample statistics make the value function overconfident on games outside them (slope 0.62 at 952K games, 0.17 at 119K) and less accurate, while reporting inflated accuracy on its own validation rows. Out-of-fold statistics remove the effect. Width, depth, dropout, weight decay, and training length each move held-out accuracy by at most 0.1pp, except training without regularization for 10 or 40 epochs ($-$0.3pp and $-$1.0pp; slope 0.57 and 0.40). Leakage dominates every capacity knob.

Search then exploits the error. With a behavioral-cloning prior and argmax root selection, raising search from 1,024 to 16,384 simulations lifts the in-sample proxy's own score from 0.754 to 0.834 and its half-B judges from 0.644 to 0.671; the out-of-fold proxy, whose own score (0.680 $\to$ 0.740) tracks the judges almost one for one, reaches 0.660 $\to$ 0.721 on the same judges and higher values on every realized index (half-B index 0.659 vs.\ 0.628, post-snapshot index 0.619 vs.\ 0.592). A small in-sample proxy (1/8 of A) shows classic over-optimization: its own score climbs to 0.941 while the half-B judges stop improving after 4,096 simulations (0.611, then 0.610). An out-of-fold proxy trained on 1/8 of the data beats an in-sample proxy with 8$\times$ the data on both realized indices (0.668 vs.\ 0.628 and 0.637 vs.\ 0.592 at 16,384 simulations).

The paper's own self-play pipeline on half A (2 seeds per cell, 300K episodes each, the behavioral-cloning bootstrap and the GD opponent trained on half A) shows the same split. Raising training simulations from 200 to 800 moves the in-sample agents' half-B judges from 0.686 to 0.607 and their degenerate rate from 27\% to 75\%, and lowers both realized indices by 0.03; the out-of-fold agents gain 0.016 under the same judges at a 3--4\% degenerate rate.

A four-seed ensemble trained on the same in-sample statistics has per-draft disagreement between 0.005 and 0.007 across the tested search budgets. Lower-confidence-bound search using that disagreement raises its mean score under the half-B judges by at most 0.012 (0.639 to 0.651), below the out-of-fold ensemble's 0.661. Held-out calibration slopes show overconfidence in the in-sample models.

\section{Calibration of the Independent References}
\label{sec:refcal}

\begin{table}[t]
\centering
\caption{Held-out accuracy and calibration slope of each reference on games none of them trained on: the paper's 38,869 test replays, and 154,387 games on the drifted 2.55.17 builds (R17 is fit on those games and omitted there).}
\label{tab:refcal}
\setlength{\tabcolsep}{3pt}
\small
\begin{tabular}{lcccc}
\toprule
& \multicolumn{2}{c}{Test (snapshot)} & \multicolumn{2}{c}{2.55.17} \\
Reference & Acc. & Slope & Acc. & Slope \\
\midrule
gN (3 OOF enriched judges) & 55.4 & 0.92 & 55.5 & 0.88 \\
gN-naive & 55.7 & 0.95 & 55.8 & 0.90 \\
RN (realized index) & 54.7 & 1.04 & 54.8 & 0.98 \\
R17 (realized index, drifted) & 53.7 & 0.71 & -- & -- \\
QM2026 & 54.0 & 0.55 & 53.6 & 0.53 \\
QM2021 & 54.7 & 0.40 & 52.6 & 0.25 \\
\midrule
Out-of-fold naive, 1 seed (in corpus) & 57.8 & 1.23 & 56.3 & 0.96 \\
\bottomrule
\end{tabular}
\end{table}

The post-snapshot judges and the realized index are close to calibrated on snapshot games. The Quick Match judges are strongly overconfident on ranked drafts (slopes 0.40--0.55): their log-odds are too extreme by a factor of about 2, so the main paper reads them for ordering, though QM2026 also enters the four-judge consensus. All references are well below the 57.8\% of an in-corpus model (the selected seed of the out-of-fold naive model; the main paper's 57.6\% is its three-seed mean), the price of independence: smaller training sets, a different time window, or a different game mode.

\section{Composition Blind Spot and Corrected References}
\label{sec:compjudge}

\paragraph{Audit} References trained on human games see few broken teams. Table~\ref{tab:compaudit} measures every reference on the 1,943,338 snapshot games, which none of them trained on, as realized minus predicted win rate per team type (no healer, no frontline, three or more of a stacking role; exclusive, in that order). Normal teams are calibrated to within 0.1pp. Degenerate teams lose 5--7pp more often than RN, gN-naive, R17, and both Quick Match judges predict; the gap survives a global slope and intercept recalibration of each judge. gN, which reads role-count features, misses 2.2pp. The post-snapshot no-drift games give the same picture with held-out predictions (gN from two-fold refits of its recipe: $-$2.8pp; RN cross-fitted: $-$5.2pp; consensus $-$4.6pp). RN's composition term shrinks each role-multiset win rate toward 50\% with a 200-game prior. In the post-snapshot games, degenerate multisets have a median of 4--10 team-games by tier. The gN-naive judge must learn role penalties from hero identities, and the QM judges learned from a mode with a different team-assembly process.

\begin{table}[t]
\centering
\caption{Realized minus predicted win rate (pp) of real teams by type, on 1.94M snapshot games no reference trained on (SE 0.2, 0.3, 1.0, 0.2pp by column). Types cover 1.37\%, 0.57\%, and 0.06\% of teams. Corr.: after the structure correction. $\Delta$LL: corrected minus uncorrected log loss ($\times 10^{-4}$).}
\label{tab:compaudit}
\setlength{\tabcolsep}{3pt}
\small
\begin{tabular}{lrrrr|rr}
\toprule
Reference & No heal. & No front. & Stack & Any & Any, corr. & $\Delta$LL \\
\midrule
gN & $-$2.6 & $-$1.4 & $-$1.2 & $-$2.2 & +0.6 & $-$0.4 \\
gN-naive & $-$5.9 & $-$3.8 & $-$9.8 & $-$5.5 & $-$0.2 & $-$2.8 \\
RN & $-$5.0 & $-$4.1 & $-$10.8 & $-$4.9 & +0.4 & $-$2.3 \\
R17 & $-$6.5 & $-$6.1 & $-$11.9 & $-$6.5 & $-$0.2 & $-$3.9 \\
QM2026 & $-$5.9 & $-$3.1 & $-$9.4 & $-$5.2 & +0.1 & $-$2.7 \\
QM2021 & $-$6.8 & $-$3.4 & $-$7.5 & $-$5.8 & +0.5 & $-$3.3 \\
Consensus & $-$4.8 & $-$3.1 & $-$7.8 & $-$4.4 & +0.2 & $-$1.9 \\
\bottomrule
\end{tabular}
\end{table}

\paragraph{Selection or cause} Weaker players might field broken teams. We joined 603,469 games (snapshot games from June 2025 on and post-snapshot games) in which all ten players have a Heroes Profile MMR recorded before the game started to those ratings and to pick order. Players on degenerate teams are rated lower (mean MMR 2,551 vs.\ 2,657 on normal teams), but matchmaking pairs them with lower-rated opponents too. Against RN, the penalty is $-$5.27pp with no controls, $-$5.18pp with the team MMR difference, $-$5.18pp adding MMR by pick position and first-pick side, and $-$4.91pp adding party counts (bootstrap SE 0.35pp). At most about 0.4pp of the 5.3pp is selection on these measures. Unmeasured differences between players remain possible.

\paragraph{Correction} For each reference $J$ we fit $p^\ast = \sigma(\mathrm{logit}\,p_J + \beta_J^\top(s_\text{own} - s_\text{opp}))$ with $s$ the three structure indicators (non-exclusive) and $p_J$ fixed, on the 260,539 post-snapshot no-drift games, using predictions $J$ makes for games it did not train on (above). The fitted no-healer, no-frontline, and stack coefficients (logits) are $-$0.10, $-$0.13, $-$0.14 (gN); $-$0.21, $-$0.20, $-$0.38 (gN-naive); $-$0.18, $-$0.25, $-$0.38 (RN); $-$0.21, $-$0.19, $-$0.36 (QM2026). The correction is antisymmetric in the two teams, so symmetrized scores stay symmetrized. On games after the 2.55.17 balance patch the penalty is smaller and the corrected references over-charge degenerate teams by about 2pp. The terms are learned from human degenerate teams, 93\% of which have exactly one defect (68\% no healer); the MCTS agents' degenerate teams have exactly one defect in 93--100\% of cases and are 80--98\% no-healer teams, the best-covered type. A structure nobody fields, such as five tanks, receives only the sum of its indicators.

\paragraph{Effect on the results} Under the corrected references, the benchmark contrasts behind sim scaling, the feature gap, training length, and the operating point change by at most 0.007 (Table~\ref{tab:compclaims}): GD opponents sampled at temperature 1 field degenerate teams too, so the correction charges both sides of the MCTS benchmark. The in-sample control contrasts sharpen by up to 0.011 because the in-sample 800-simulation agent fields 27\% degenerate teams. In the twelve-strategy tournament the constrained agents gain (constrained vs.\ unconstrained MCTS 0.497 $\to$ 0.513, masked vs.\ unmasked greedy 0.500 $\to$ 0.511, $\pm$0.004 and $\pm$0.003). Consensus means move from 0.579 to 0.589 for constrained MCTS (wins 19 $\to$ 22), 0.584 to 0.583 for K (22 $\to$ 19), 0.514 to 0.494 for the Gourdeau estimator (46\% degenerate), and 0.480 to 0.455 for TCQL (46\%). Within the top six, rank agreement of the Quick Match judges with the consensus falls from 1.00 to 0.83, because the correction moves constrained MCTS ahead of K and the Gourdeau estimator behind both greedy agents, orderings those judges do not share.

\begin{table}[t]
\centering
\caption{Paper claims under the uncorrected and corrected references. MCTS benchmark contrasts, root temperature 1 (5 seeds; 3 for E and the in-sample rows); $\pm$: seed SE of the corrected values.}
\label{tab:compclaims}
\setlength{\tabcolsep}{2.5pt}
\small
\begin{tabular}{lrrrr}
\toprule
& \multicolumn{2}{c}{gN} & \multicolumn{2}{c}{RN} \\
Contrast & Unc. & Corr. & Unc. & Corr. \\
\midrule
F $-$ B & $-$.005 & $-$.003 & $-$.005 & $-$.001 $\pm$.006 \\
J $-$ F & +.003 & +.003 & +.001 & +.001 $\pm$.006 \\
F $-$ B, in-sample & $-$.004 & $-$.004 & $-$.002 & $-$.002 $\pm$.005 \\
J $-$ F, in-sample & $-$.012 & $-$.016 & $-$.006 & $-$.013 $\pm$.006 \\
J in-sample $-$ J & $-$.023 & $-$.029 & $-$.012 & $-$.023 $\pm$.006 \\
F $-$ K & $-$.001 & +.002 & $-$.008 & $-$.002 $\pm$.006 \\
E $-$ B & $-$.007 & $-$.006 & $-$.009 & $-$.007 $\pm$.004 \\
F $-$ E & +.001 & +.003 & +.003 & +.007 $\pm$.004 \\
F, argmax $-$ T=1 & +.002 & +.002 & +.002 & +.002 $\pm$.001 \\
\bottomrule
\end{tabular}
\end{table}

\paragraph{Composition table of gN} gN reads a role-composition table for 4 of its 283 inputs. Any table built from the community aggregates overlaps our snapshot, so gN builds its own from its post-snapshot games (out of fold for its training rows) and shares no statistic with any agent. It reaches 55.4\% test accuracy with slope 0.92.

\section{Judge Attenuation}
\label{sec:atten}

An independent judge is a noisier model than the value function an agent optimized, so a real gain shrinks under it. As a benchmark for that shrinkage we regress each judge's prediction on the value function's prediction over the 38,869 test replays, which neither trained on (Table~\ref{tab:atten}). With no over-optimization, a gain $d$ under the value function should appear as about slope $\times\, d$ under the judge, if the value function's errors on agent drafts resemble its errors on human drafts.

\begin{table}[t]
\centering
\caption{Slope of each composition-corrected judge on the value function's prediction, test replays (bootstrap SE $\le$ 0.01).}
\label{tab:atten}
\small
\setlength{\tabcolsep}{3pt}
\begin{tabular}{lccccc}
\toprule
Value function & gN & gN-nv. & RN & QM26 & Cons. \\
\midrule
Out-of-fold enriched & 0.59 & 0.66 & 0.45 & 0.49 & 0.55 \\
\bottomrule
\end{tabular}
\end{table}

Against this benchmark, neither doubling of training simulations transmits in the out-of-fold pipeline (200 $\to$ 400: own +0.017, predicted gN +0.010, observed $-$0.003 $\pm$ 0.004; 400 $\to$ 800: own +0.011, predicted +0.006, observed +0.003 $\pm$ 0.004). With in-sample statistics the own score rises faster (+0.043, then +0.026) while gN falls ($-$0.004 $\pm$ 0.004, then $-$0.016 $\pm$ 0.005): over-optimization.

\section{Feature-Group Sweep (Fractional Factorial)}
\label{sec:sweep}

We measure each feature group's effect with out-of-fold statistics using a $2^{10-3}$ fractional factorial: base factors map\_type, role\_counts, hero\_wr, team\_avg\_wr, hero\_map\_wr, map\_delta, pairwise\_counters (A--G), and generated factors pairwise\_synergies $=$ ABCD, counter\_detail $=$ ABEF, synergy\_detail $=$ ACEG. Every word of the defining relation has length at least 5 (resolution V), so main effects are unaliased with each other and with all two-factor interactions, and each main effect equals the full-factorial marginal up to four-factor and higher interactions. Each of the 128 configurations trains one 256$\to$128 model (seed 42, learning rate $5\times10^{-4}$, validation early stopping). Effect SE, from the residual after fitting main effects: 0.02pp (test accuracy), 0.01pp (post-snapshot accuracy), $4\times10^{-5}$ (post-snapshot log-loss).

\begin{table}[t]
\centering
\caption{Main effect of each feature group (with minus without), out-of-fold statistics.}
\label{tab:sweep}
\setlength{\tabcolsep}{2.5pt}
\small
\begin{tabular}{lccc}
\toprule
Group & Test & Post & Post LL$\times10^{3}$ \\
\midrule
role\_counts & +0.11 & +0.13 & $-$0.72 \\
pairwise\_counters & +0.04 & +0.03 & +0.08 \\
pairwise\_synergies & +0.03 & +0.02 & +0.01 \\
map\_type & +0.02 & +0.01 & $-$0.02 \\
map\_delta & $-$0.00 & +0.02 & +0.04 \\
synergy\_detail & $-$0.00 & $-$0.05 & +0.36 \\
team\_avg\_wr & $-$0.03 & $-$0.04 & +0.14 \\
counter\_detail & $-$0.07 & $-$0.08 & +0.69 \\
hero\_wr & $-$0.08 & $-$0.11 & +0.41 \\
hero\_map\_wr & $-$0.09 & $-$0.10 & +0.47 \\
\bottomrule
\end{tabular}
\end{table}

Role counts, which read no statistics, have the largest effect. Every effect is below 0.15pp in absolute value on both held-out sets.

\section{Added Strategies and the Opponent-Swap Test}
\label{sec:added}

\paragraph{Masked win-rate drafter} At each of our picks the drafter takes the available hero with the highest win rate on the map and tier, after the structural mask of the main paper's constrained agents. Win rates are the deploy statistics the greedy agents read; each hero's rate on a map and tier is shrunk toward its tier-wide rate with 500 pseudo-games. A ban removes the highest-ranked available hero. The rule was fixed before any tournament score. It reads no value function and no hero pair, so it cannot react to the opponent. Over its 22 ordered pairs it fields no degenerate team (synergy +0.24). Composition-corrected means are 0.588 under gN (19 pairs won), 0.591 under RN (21), 0.590 under gN-naive, 0.580 under R17, 0.542 under QM2026 and 0.569 under QM2021, so its consensus mean is 0.578 (16 wins). Constrained MCTS has 0.590 and 0.574 under gN and RN and 0.605 under QM2026. The gap between the two agents is almost entirely the Quick Match judges.

\paragraph{Masked K} The featureless MCTS agent K pools its five seeds (draft $i$ uses seed $i \bmod 5$) and takes the policy argmax under the same structural mask as constrained MCTS, so it differs from constrained MCTS only in the value function's features. Against the eleven learned strategies its composition-corrected means are 0.606 (gN), 0.593 (RN), 0.601 (gN-naive), 0.604 (R17), 0.619 (QM2026) and 0.665 (QM2021); consensus 0.605, 21 of 22 pairs won. It fields no degenerate team, a healer in every draft, 41 distinct heroes at 3.89 bits of pick entropy, synergy +0.79 and counter $-$0.41. Head to head it ties constrained MCTS (consensus 0.501 $\pm$ 0.002; per judge 0.494 to 0.508 except QM2021, 0.519) and beats unmasked K (0.518 $\pm$ 0.003) and unmasked MCTS (0.510).

\paragraph{Opponent-swap protocol} We draft 1,000 states with GD (temperature 1) up to one of our picks at steps 7, 8, 13 and 14 (about 250 each), cycling through maps and tiers. For each state a second state keeps the map, tier, bans and our picks and resamples the opponent's picks from GD. Each policy picks once in each state. A pair counts only if neither pick is taken in the other state, so both picks were available in both; 706 to 958 of the 1,000 pairs count, depending on the policy. For pick $x$ against opponent team $O$, $c(x, O)$ is the counter score of Section~\ref{sec:bench}. The crossover score is $\tfrac12[c(x_1, O_1) - c(x_2, O_1) + c(x_2, O_2) - c(x_1, O_2)]$; it is zero for a policy that ignores the opponent and positive when each pick counters the team it was made against. Table~\ref{tab:oppswap} lists every tournament strategy.

\begin{table}[h]
\centering
\caption{Opponent-swap test, 1,000 matched states. Changed: share of counted pairs in which the pick differs between the two states. Crossover: mean counter-score crossover (SE over pairs); per change: crossover over the pairs whose pick changed.}
\label{tab:oppswap}
\footnotesize
\setlength{\tabcolsep}{3pt}
\begin{tabular}{lcccc}
\toprule
Strategy & Pairs & Changed & Crossover & Per change \\
\midrule
Constrained greedy & 941 & 0.90 & +0.41 (0.03) & +0.45 \\
Enriched greedy & 958 & 0.90 & +0.39 (0.03) & +0.43 \\
MCTS & 895 & 0.31 & +0.25 (0.02) & +0.81 \\
Constrained MCTS & 888 & 0.29 & +0.24 (0.02) & +0.81 \\
K & 868 & 0.29 & +0.19 (0.02) & +0.65 \\
Masked K & 856 & 0.27 & +0.18 (0.02) & +0.65 \\
CQL (naive) & 712 & 0.42 & +0.06 (0.02) & +0.16 \\
GD & 706 & 0.42 & +0.05 (0.02) & +0.13 \\
CQL (enriched) & 722 & 0.39 & +0.05 (0.01) & +0.12 \\
G\&A discrim. & 908 & 0.67 & +0.02 (0.02) & +0.03 \\
Gourdeau est. & 929 & 0.81 & $-$0.01 (0.03) & $-$0.01 \\
TCQL & 897 & 0 & 0 & -- \\
Masked win-rate & 882 & 0 & 0 & -- \\
\bottomrule
\end{tabular}
\end{table}

The greedy searchers change almost every pick, because their value functions read the opponent's heroes directly, and they collect the largest crossover. The MCTS agents change fewer picks, but each change is a strong counter: about 0.8 per changed pick, against 0.45 for greedy search. GD and the CQL policies change about two picks in five with little counter gain, which looks like responding to the opponent's heroes the way human players do on average, not to their weaknesses. The Gourdeau estimator and the G\&A discriminator change picks without countering. TCQL and the win-rate drafter never change a pick.

\section{Offline RL: Sweeps and Robustness}
\label{sec:cqlrobust}

Table~\ref{tab:rich} shows selected cells from the offline-RL sweeps. The full sweeps, trained on our split with naive (identity) features unless noted and selected on validation games, all end either at a policy anchored to observed play or at an unanchored one that fields broken teams.

\textit{Anchored families.} CQL across $\alpha \in \{0.1, 0.5, 1, 2, 5\}$ (5 seeds $\times$ 1000 drafts each): 99.2--99.8\% healer and 0.8--3.2\% degenerate, best at $\alpha = 2$ (99.6\%, 0.8\%). BC-regularized CQL across $\beta \in \{0.1, 0.5, 1, 2\}$: 100\% healer, 1.0--3.5\% degenerate, GD agreement 73--74\%, synergy $-$0.03 to +0.01. Sweep cells here are 200 drafts with one opponent-sampling seed, and their synergy runs about 0.1 above the 5 $\times$ 1000-draft evaluations of Table~IV (BC-CQL $\beta = 1$: $-$0.01 here, $-$0.13 there); degenerate rates in 200 drafts carry about $\pm$1.5pp of sampling error. IQL across $\tau \in \{0.7, 0.8, 0.9\} \times \beta \in \{1, 3\}$ (200 drafts each): 99.5--100\% healer, 1.5--4.0\% degenerate, GD agreement 67--73\%, statistically indistinguishable cells. Every anchored method matches the logged human action as often as the behavioral-cloning drafter (top-1 13.3--13.5\% against GD's 13.0--13.2\%). The bootstrapped variants overestimate: CQL with temporal-difference targets has a held-out mean Q of 0.87--0.89 against a base rate of 0.5 (0.78 at $\alpha = 0.1$), although its policy still plays like humans, and IQL with temporal-difference targets at temperature 30 has a policy loss that rises during training, with a Q that is uninformative at every setting (held-out log loss 0.68--0.69).

\textit{The outcome signal.} On held-out transitions (rows whose logged action is masked excluded), the unpenalized Q-network ($\alpha = 0$) has log loss 0.6922 and AUC 0.524 over all steps, 0.6894 and 0.548 at the final pick. CQL's penalty raises held-out log loss by pushing up the logged action's value (0.761 at $\alpha = 0.5$, 0.786 at $\alpha = 1$) and lowers AUC to 0.508 and 0.506.

\textit{Unanchored families.} The unpenalized control ($\alpha = 0$; 5 $\times$ 1000): 84.5\% healer, 27.6\% degenerate, 85 heroes, GD agreement 4.6\%, synergy +0.35. MCQ across $\lambda \in \{0.5, 0.7, 0.9\}$ (5 $\times$ 1000 each): held-out loss 0.69336, 0.69338, and 0.69408, so the tournament uses $\lambda = 0.5$; healer 92.3\%, 86.2\%, 80.2\%; degenerate 24.3\%, 25.3\%, 25.6\%; 83, 86, and 90 heroes; GD agreement 4.1--4.9\%; synergy $-$0.20, $-$0.02, +0.08. TCQL, a thresholded variant in the spirit of MCQ, replaces the CQL penalty with $\mathbb{E}_s[\sum_a (Q(s,a) - \tau)_+] - \mathbb{E}_\mathcal{D}[(Q(s,a) - \tau)_+]$ on the sigmoid Q-values (weight 1), so only values above $\tau$ are pushed down; across $\tau \in \{0.4, \dots, 0.8\}$ (200 drafts each): 81.5--90.0\% healer, 19.5--31.5\% degenerate, GD agreement 2.9--6.6\% (the 5$\times$1000-draft evaluation of $\tau = 0.5$ in the main paper, 34.0\% degenerate, supersedes its 200-draft cell); at $\tau = 0.4$ it narrows to 12 heroes (synergy $-$0.61). A threshold sits poorly on sigmoid-bounded discrete Q-values: on 200,000 held-out states the $\tau = 0.5$ network (the tournament's TCQL) has 122 of 256 units of hidden layer 2 and 52 of 128 of layer 3 that never activate, and the $\tau = 0.6$ network 47 and 31, against 21 and 16 for a CQL $\alpha = 1$ control.

\textit{Capacity robustness.} At $\alpha = 1$ we trained three network sizes (256$\to$128$\to$64, 512$\to$256$\to$128, 1024$\to$512$\to$256), one network each, each evaluated in five independent draws of 200 drafts. Across all 15 evaluations the anchoring signals are stable: healer 99.5--100\%, degenerate 0.0--4.0\%, 50--59 distinct heroes, entropy 4.50--4.68 bits, GD agreement 65.6--75.3\%, synergy $-$0.07 to +0.12. No configuration is interaction-aware.

\textit{CQL with enriched features.} Trained on out-of-fold enriched transition features and evaluated with statistics over the whole training split, enriched CQL behaves like naive CQL: 2.6--3.6\% degenerate, synergy $-$0.13 to $-$0.17, entropy 4.52--4.58 bits, GD agreement 68.6--72.0\%. Under gN and RN these agents score 0.50--0.51 against behavioral-cloning opponents.

\section{Benchmark, Composition, and Tournament Results}
\label{sec:bench}

This section collects the drafting benchmark behind the main paper's setup: composition safety, interaction metrics, the MCTS benchmark, and the twelve-strategy tournament.

\paragraph{Composition safety} Healer rate and degenerate rate. A team is degenerate if it has no healer, no frontline hero (a tank or bruiser; Uther counts as frontline when the team has another healer), or three or more heroes of a stacking role (tank, melee assassin, support, or healer; three ranged assassins or three bruisers are allowed). These are the classes that role-composition statistics show underperforming (the 51,090 no-healer teams in our training split win 42.4\% of their games) or absent from play.

\paragraph{Interaction-awareness} Pairwise deltas (in percentage points) are observed win rates minus an additive hero-strength baseline: the counter delta of $a$ against $b$ is $\mathrm{WR}(a\text{ vs }b) - [\mathrm{WR}(a) + (100 - \mathrm{WR}(b)) - 50]$, and the synergy delta of teammates is $\mathrm{WR}(a\text{ with }b) - [\mathrm{WR}(a) + \mathrm{WR}(b) - 50]$. \textit{Counter responsiveness} averages our heroes' counter deltas against the opponents, \textit{synergy} averages within-team synergy deltas. Behavioral signals need no statistics: distinct heroes, pick entropy, top-10 concentration, and agreement with the behavioral-cloning model (GD similarity: the majority-vote pick of the five GD models). The interaction metrics use the community aggregates as an evaluation instrument; no value function in this paper reads them. The two sources overlap in games, but the metric does not merely echo the features: the featureless MCTS agent K (the main paper), whose value function reads no statistics, reaches the highest tournament synergy of any agent (+0.95, Table~\ref{tab:tourn}). On 260,539 post-snapshot games, which the aggregates cannot contain, both metrics predict outcomes monotonically after controlling for team strength (+8.8pp synergy, +9.2pp counter, top vs.\ bottom decile; supplement).

\begin{table*}[t]
\centering
\caption{Rich evaluation (5 opponent-sampling seeds $\times$ 1000 drafts per strategy). Counter and synergy use the community aggregates for every row. G\&A: Gourdeau and Archambault. Ent.: pick entropy (bits); T10: top-10 hero share; GD: agreement with the majority-vote pick of the five GD models. Every row is trained on our split; supervised and offline-RL rows are selected on validation games, MCTS checkpoints as in the main paper. The anchored rows read no aggregate statistics except CQL (enr.), trained on out-of-fold features and run with statistics over the whole training split; greedy and MCTS rows use out-of-fold value functions. Greedy rows search the 256$\to$128 value functions of the main paper (or Gourdeau's) against GD opponents. MCTS rows: the five F seeds (the main paper), policy argmax, or with 200 simulations of search at each move.}
\label{tab:rich}
\begin{tabular}{lcccccccc}
\toprule
Strategy & Heal\% & Deg.\% & Counter & Synergy & Dist. & Ent. & T10\% & GD\% \\
\midrule
\multicolumn{9}{l}{\textit{Anchored to observed play}} \\
GD baseline & 99.4 & 1.3 & $-$.02 & $-$.16 & 66 & 4.50 & 71.7 & 81.4 \\
G\&A discriminator & 99.9 & 0.3 & $-$.02 & $-$.06 & 87 & 4.39 & 75.3 & 40.6 \\
CQL $\alpha$=.5 (naive) & 99.2 & 1.1 & $-$.01 & $-$.09 & 70 & 4.72 & 66.5 & 66.6 \\
CQL $\alpha$=1 (naive) & 99.6 & 0.9 & $-$.01 & $-$.14 & 69 & 4.65 & 68.1 & 72.7 \\
CQL $\alpha$=.5 (enr.) & 99.6 & 2.6 & $-$.01 & $-$.13 & 71 & 4.52 & 71.3 & 68.6 \\
CQL $\alpha$=2 (enr.) & 99.9 & 3.6 & $-$.01 & $-$.17 & 73 & 4.58 & 69.6 & 72.0 \\
CQL BC form $\beta$=1 & 99.8 & 1.6 & $-$.01 & $-$.13 & 71 & 4.65 & 68.3 & 73.2 \\
IQL $\tau$=.9 $\beta$=3 & 99.7 & 3.2 & $-$.01 & $-$.13 & 73 & 4.72 & 66.9 & 67.6 \\
\midrule
\multicolumn{9}{l}{\textit{Mild or no pessimism}} \\
TCQL $\tau$=.5 & 84.1 & 34.0 & $-$.04 & $-$.18 & 69 & 4.54 & 66.9 & 5.6 \\
MCQ $\lambda$=.5 & 92.3 & 24.3 & $-$.04 & $-$.20 & 83 & 5.12 & 55.5 & 4.7 \\
No penalty ($\alpha$=0) & 84.5 & 27.6 & $-$.06 & $+$.35 & 85 & 5.47 & 45.5 & 4.6 \\
\midrule
\multicolumn{9}{l}{\textit{Value search}} \\
Gourdeau estimator (greedy) & 66.8 & 40.1 & $-$.08 & $+$.36 & 90 & 5.73 & 43.2 & 4.9 \\
Enriched (greedy) & 87.2 & 18.6 & $+$.09 & $+$.36 & 90 & 6.01 & 33.7 & 4.8 \\
MCTS & 96.8 & 3.8 & $+$.06 & $+$.86 & 35 & 3.68 & 85.6 & 3.4 \\
MCTS + search at play & 96.5 & 4.1 & $+$.06 & $+$.85 & 34 & 3.68 & 85.2 & 3.7 \\
Constrained MCTS & 100 & 0.0 & $+$.04 & $+$.71 & 40 & 3.70 & 86.0 & 4.0 \\
\bottomrule
\end{tabular}
\end{table*}

\begin{table}[t]
\centering
\caption{MCTS benchmark (5 seeds, E and in-sample rows 3 seeds, $\times$ 1,000 drafts; 200 simulations at inference, root temperature 1). Train: the value function the configuration optimized (K: the hero-identity model). gN, RN, QM26: composition-corrected references (the main paper); gN seed SE 0.002--0.003. Syn.\ is measured with search and root temperature 1 against the GD pool; the MCTS rows of Tables~\ref{tab:rich} and~\ref{tab:tourn} play the policy argmax or search with an argmax root, so their values are lower.}
\label{tab:mcts}
\setlength{\tabcolsep}{2.8pt}
\small
\begin{tabular}{lrccccrc}
\toprule
Config & Sims & Train & gN & RN & QM26 & Deg\% & Syn. \\
\midrule
\multicolumn{8}{l}{\textit{Out-of-fold value function}} \\
B & 200 & .723 & .665 & .641 & .664 & 10.5 & +1.06 \\
F & 400 & .740 & .662 & .641 & .669 & 4.5 & +1.29 \\
J & 800 & .751 & .665 & .642 & .668 & 7.9 & +1.31 \\
E (1M ep.) & 200 & .726 & .659 & .634 & .656 & 9.2 & +1.21 \\
K (no feat.) & 400 & .726 & .660 & .643 & .672 & 15.4 & +1.24 \\
\midrule
\multicolumn{8}{l}{\textit{In-sample statistics (control)}} \\
B & 200 & .742 & .656 & .633 & .648 & 11.4 & +1.07 \\
F & 400 & .785 & .653 & .631 & .665 & 11.7 & +1.56 \\
J & 800 & .811 & .636 & .619 & .671 & 27.1 & +2.02 \\
\bottomrule
\end{tabular}
\end{table}

\paragraph{Simulation scaling} Under its own value function the out-of-fold agent gains with every doubling of training simulations, 0.723 $\to$ 0.740 $\to$ 0.751 from 200 to 800 (+0.028 $\pm$ 0.005). On the test replays, the regression slope of gN predictions on the out-of-fold value function's is 0.59 (consensus 0.55; supplement). That predicts about +0.016 for this step if the value function errs on agent drafts as it does on human drafts. Measured, the independent references stay flat (gN +0.000 $\pm$ 0.004, RN +0.000, consensus +0.003).

The in-sample control shows what statistics containing each game's outcome do to the same pipeline. At matched simulations the in-sample value function raises the agents' own score (+0.020, +0.045, and +0.060 at 200, 400, and 800 simulations) and lowers gN and RN (at 800: gN $-$0.029 $\pm$ 0.006, RN $-$0.023 $\pm$ 0.006, consensus $-$0.015 $\pm$ 0.006). The gap widens with search (from 200 to 800 simulations the own score rises 0.069 while gN falls 0.020 $\pm$ 0.006, less than in the half-split control of the main paper, whose value functions train on half the data); QM2026 alone does not penalize these agents. A value function calibrated on human games is not thereby calibrated on the drafts search produces.

\paragraph{Training length and root selection} Training for 1M episodes instead of 300K does not help (E $-$ B: $-$0.005 $\pm$ 0.003 under the consensus). An argmax root at inference adds 0.000--0.004 gN over temperature 1. Our operating point is F. It scores within noise of J under every reference with half the training simulations and has the lowest benchmark degenerate rate (4.0\% with an argmax root); as the policy argmax without search, which is how it plays in the tournament, it fields 3.8\% in Table~\ref{tab:rich}.

\paragraph{Features} Under the independent references at matched search, we find no resolved difference in win probability between the enriched agent and the featureless one: F $-$ K is +0.002 ($\pm$0.004) under gN, $-$0.002 ($\pm$0.006) under RN, $-$0.002 ($\pm$0.004) under QM2026, and $-$0.001 ($\pm$0.003) under the consensus. What the features buy is composition safety. The featureless agent fields 15.4\% degenerate teams against 4.5\% ($-$10.9 $\pm$ 3.3pp across seeds; healer rate $+$10.0 $\pm$ 3.1pp): a value function that reads no role statistics must infer from hero identities alone that a team lacks a healer, and the training data rarely show one. Under the structural mask the two agents tie (the main paper).

\begin{table}[t]
\centering
\caption{Tournament: mean win probability over each strategy's 22 ordered pairs, and pairs won under the consensus (mean of gN, gN-naive, RN, QM2026; SE 0.001). All judges composition-corrected (the main paper). Deg.: degenerate rate of the strategy's teams; it differs from Table~\ref{tab:rich} because the opponents differ. $^\dagger$Plays the eleven learned strategies, not the win-rate drafter.}
\label{tab:tourn}
\setlength{\tabcolsep}{1.8pt}
\footnotesize
\begin{tabular}{lcccccccc}
\toprule
Strategy & Cons. & Won & gN & RN & QM26 & QM21 & Deg. & Syn. \\
\midrule
Constr.\ MCTS & .589 & 22 & .590 & .574 & .605 & .643 & 0.0 & +.76 \\
MCTS, no feat. & .583 & 19 & .567 & .563 & .617 & .651 & 21.6 & +.95 \\
MCTS & .583 & 19 & .570 & .563 & .613 & .644 & 11.5 & +.89 \\
Top win rate & .578 & 16 & .588 & .591 & .542 & .569 & 0.0 & +.24 \\
Constr.\ greedy & .516 & 14 & .530 & .521 & .485 & .448 & 0.0 & +.50 \\
Enr.\ greedy & .505 & 11 & .514 & .507 & .478 & .444 & 21.6 & +.55 \\
Gourdeau est. & .494 & 11 & .496 & .487 & .484 & .464 & 45.7 & +.62 \\
TCQL & .455 & 8 & .458 & .469 & .446 & .490 & 46.3 & +.09 \\
CQL naive & .431 & 3 & .428 & .437 & .436 & .431 & 2.5 & $-$.23 \\
CQL enr. & .426 & 3 & .425 & .432 & .433 & .418 & 3.4 & $-$.22 \\
GD & .420 & 3 & .417 & .427 & .432 & .405 & 3.8 & $-$.24 \\
G\&A discrim. & .420 & 3 & .418 & .428 & .429 & .395 & 0.7 & $-$.10 \\
\midrule
MCTS + search$^\dagger$ & .582 & 19 & .571 & .566 & .608 & .647 & 12.9 & +.87 \\
MCQ $\lambda$=.5$^\dagger$ & .497 & 10 & .486 & .510 & .481 & .516 & 33.6 & $-$.37 \\
Constr.\ MCTS, no feat.$^\dagger$ & .605 & 21 & .606 & .593 & .619 & .665 & 0.0 & +.79 \\
\bottomrule
\end{tabular}
\end{table}

Among the twelve round-robin entrants (Table~\ref{tab:tourn}), MCTS agents and the win-rate drafter lead (gN 0.57--0.59, RN 0.56--0.59), greedy searchers follow (gN 0.50--0.53, RN 0.49--0.52), TCQL is at or below the weakest of them (0.46--0.47), and behavior-anchored methods trail (gN 0.42--0.43, RN 0.43--0.44). Constrained MCTS wins all 22 of its ordered pairs under the consensus, gN, and gN-naive, and 21 under RN, where it splits the two seatings with the drafter; under the uncorrected references, which under-charge broken teams, it wins 19 and K 22. The rank correlation of each reference with the mean of the other consensus references is 0.92--0.97. Within the top six they agree less (0.43--0.83): the Quick Match judges place the greedy agents below 0.5 and K first, and gN and RN place the drafter at or near the top, so the finer ordering among the leaders is open. MCTS and K tie head to head (0.500 $\pm$ 0.003), as in the benchmark, while K fields nearly twice as many degenerate teams. Constrained MCTS beats K head to head (0.511 $\pm$ 0.003). MCQ wins 10 of 22 ordered pairs and loses head to head to every MCTS agent (0.395--0.403). Search at play changes 2\% of the policy's picks and scores 0.504 $\pm$ 0.004 head to head against the policy argmax, with a similar degenerate rate against the ten opponents they share (10.8\% against 11.7\%).

\paragraph{Hero strength} The masked win-rate drafter fields no degenerate team and beats both greedy searchers head to head (0.545 against constrained greedy, 0.556 against enriched greedy) and the Gourdeau estimator (0.560). Masked hero-strength selection beats every greedy searcher and reaches tournament averages comparable to MCTS under gN and RN. Head to head under the consensus every MCTS agent beats it (constrained MCTS 0.531 $\pm$ 0.004, MCTS 0.525, K 0.516), mostly through QM2026 (0.58--0.59). Constrained MCTS scores 0.520 $\pm$ 0.004 against it under gN and 0.497 under RN; across the MCTS agents and these two judges, scores range from 0.480 to 0.520.

\paragraph{Features under the mask} Masked K scores 0.605 against the eleven learned strategies, wins 21 of 22 pairs, and ties constrained MCTS head to head (0.501 $\pm$ 0.002; gN 0.500, RN 0.503); it beats unmasked K 0.518 $\pm$ 0.003. Once the mask provides safety by rule, we measure no win-probability gain from the features.

\section{Training Seeds of the Floor-Trained Policy}

We trained eight seeds of the ordinary MCTS policy and eight of the floor-trained policy, each with the same configuration and budget (400 training simulations, 300,000 episodes), and scored every seed's drafts in the same 1,000 family-a worlds at default dials against the same opponents. Table~\ref{tab:s-seeds} gives seed-level contrasts at default dials and with every below-floor team 2 points worse than the model believes: the difference of mean true strength over seeds, with 95\% confidence intervals over seeds (Welch for contrasts between policies, paired by seed for the floor's cost on one policy).

\begin{table}[h]
\centering
\caption{Seed-level contrasts in true win probability (points), family a, 8 seeds per arm.}
\label{tab:s-seeds}
\begin{tabular}{lcc}
\toprule
Contrast & Mean (95\%) & Range by seed \\
\midrule
Floor's cost, ordinary policy & 2.6 (2.1, 3.2) & 1.6 to 3.5 \\
Floor's cost, floor-trained policy & 1.1 (0.9, 1.3) & 0.8 to 1.5 \\
Floor-trained + floor vs ordinary & +0.5 ($-$0.8, 1.8) & \\
Floor-trained + floor + search vs ordinary & +0.8 ($-$0.5, 2.0) & \\
Floor-trained vs ordinary, no floor & +1.6 (0.4, 2.8) & $-$0.7 to 3.1 \\
\midrule
\multicolumn{3}{l}{\textit{Every below-floor team 2 points worse (flat grade, $\delta = -2$)}} \\
Floor's cost, ordinary policy & 1.5 (1.0, 2.0) & 0.5 to 2.2 \\
Floor's cost, floor-trained policy & 0.3 (0.2, 0.5) & 0.1 to 0.6 \\
Floor-trained + floor vs ordinary & +1.7 (0.4, 3.1) & \\
Floor-trained + floor + search vs ordinary & +2.0 (0.7, 3.3) & \\
\bottomrule
\end{tabular}
\end{table}

The seed-to-seed SD of true strength is 1.1 for the ordinary policy and 1.2 for the floor-trained one. The flat grade gives the same values at default dials, where it and the graded rule agree.

\section{Expert Study}
\label{sec:expertsupp}

Every judge in this paper is a model. We tested whether experienced players could provide a check that is not a model, in a study registered before data collection (\osfurl). The agents were the tournament strategies other than K and the win-rate drafter, retrained with the same recipe on the ranked games before 2026-09-01; the MCTS agent is F at 400 simulations. The 901 items were 280 pairs of drafts from a tournament among these agents, 570 real ranked games from 2026-09-14 to 09-27 with known winners (190 per tier), 40 shared calibration games, and 11 checks that pair a real draft with a role-broken team (8 screening items and 3 late attention checks). Fourteen paid raters each had a fixed set of 240 items, which assigned every machine pair to three raters. For each item a rater picked the better draft, gave a win probability, and stated a confidence, without being told which drafts were real. A machine pair's model label is the mean of three draft-only value functions (hero identity, hero strength, enriched) trained out of fold on the same snapshot and composition-corrected as in the main paper, so it is not one of the tournament's references; pairs within 0.02 of 0.5 are near-ties and excluded.

Eleven raters completed all 240 items. Collection closed on 2026-10-26, three weeks before the registered six-week limit, with outcome results unexamined. Nine raters passed the registered gates (at least 7 of 8 screening items, 2 of 3 late attention checks, and a median response time of at least 10 seconds); two failed the screener; including them moves no estimate by more than 0.03 (exploratory). All nine analyzed raters play at Master rank by their public ranked match histories; three were chosen in advance for long experience with low-rank play. Three raters who stopped early are excluded (supplement). Estimates come from mixed-effects logistic models with rater and item random effects, except the low-tier real-game fit, which keeps only the item effect (Table~\ref{tab:expert}; supplement).

\begin{table}[t]
\centering
\caption{Expert study (9 raters). Top: agreement, the share of judgments on model-generated pairs that pick the draft the model label prefers. Bottom: the share that pick the winner of a real game; the model column is the label's accuracy on the same 190 games per tier (descriptive, post hoc; Wilson 95\% CI). Expert estimates are mixed-model estimates with 95\% CIs; counts are items / judgments.}
\label{tab:expert}
\setlength{\tabcolsep}{2.5pt}
\small
\begin{tabular}{lccc}
\toprule
 & Items / judg. & Experts & Model label \\
\midrule
\multicolumn{4}{l}{\textit{Agreement with the model on generated pairs}} \\
All clear pairs & 255 / 488 & .50 [.43, .57] & -- \\
Anchored vs.\ searching & 113 / 215 & .33 [.21, .48] & -- \\
\midrule
\multicolumn{4}{l}{\textit{Picking the winner of real games}} \\
Low tier & 190 / 387 & .60 [.54, .65] & .60 [.53, .67] \\
Mid tier & 190 / 387 & .49 [.43, .55] & .54 [.47, .61] \\
High tier & 190 / 387 & .53 [.46, .59] & .56 [.49, .63] \\
\bottomrule
\end{tabular}
\end{table}

\textit{Can experts check the models?} On the 570 real games, estimated expert accuracy was 60\% at low tier (Holm-adjusted $p = 0.004$), 49\% at mid tier, and 53\% at high tier; pooled accuracy was 53\%. Draft-only models do only somewhat better: our value functions reach 57--58\% on held-out games, and on these same games the model label scores 57\% overall (60\%, 54\%, 56\% by tier; the enriched value function alone 59\%). With 190 games per tier those model figures carry intervals of about $\pm$7pp, so the tiers do not separate experts from models; above the lowest tier the experts are indistinguishable from chance (95\% CIs 0.43--0.55 and 0.46--0.59). Their win probabilities did not track the model's (slope 0.00, 95\% CI $-$0.22 to 0.22), and raters chosen in advance for low-rank experience were no more accurate at low tier (57\% vs.\ 59\%).

\textit{Agreement on generated pairs.} Experts picked the model-preferred draft in 244 of 488 judgments (0.50, 95\% CI 0.43--0.57), so the registered primary test finds no above-chance agreement. The result does not move with the near-tie threshold (0.49--0.52), with any single value function as the label (0.50 each), or without the seven heroes the following patch (2.57) changed (0.49). Where a behavior-anchored strategy (GD, either CQL, or the discriminator) met a searching one, experts chose the anchored draft in 152 of 215 judgments. The label itself preferred the anchored draft in some of these pairs, so pooled agreement with it is 80 of 215 (mixed-model estimate 0.33, 0.21--0.48, $p = 0.03$). On the other pairs they agreed with the model 57--66\% of the time (exploratory). A Bradley--Terry ranking of the ten strategies from their choices does not track the tournament order (Spearman $\rho = -0.19$, one-sided $p = 0.71$). Agreement did not fall on items the value-function ensemble found unfamiliar (slope $+$0.24 log-odds per SD of ensemble variance, CI $-$0.02 to 0.50), which meets the registered robustness criterion.

\textit{What it means.} The experts preferred human-like drafts, but above the lowest tier their picks of real winners were indistinguishable from chance, though the intervals also allow moderate skill. The study did not establish expert judgment as an independent check on the model-based ranking. As registered, the evaluation claims of this paper rest on outcome-based evidence: held-out accuracy and calibration, the composition audit on real games, and the independent references, which share no data with any agent but are themselves models.
\section{Expert Study: Design and Analysis Notes}

\paragraph{Registration and materials} The plan was registered before data collection (\osfurl). The item pool, analysis code, frozen ratings export, and outputs are in the repository.

\paragraph{Raters and gates} Fourteen slots, each a fixed sequence of 240 items: 8 screening items (a real winning draft against a role-broken team) and 40 calibration games shared by all raters in positions 1--48, then 60 machine pairs and 129 real games, with 3 attention checks at positions 121, 181, and 231. Eleven raters completed 240 items. Gates: at least 7 of 8 screening items, at least 2 of 3 attention checks, and a median response time of at least 10 seconds. Two completers failed the screener (6 of 8 each); nine passed every gate. Three raters stopped at 7, 24, and 63 items. Gates are applied as written, so a rater who never reached a gate's items cannot pass it and all three are excluded; holding them only to the items they reached admits the rater with 63 items; no conclusion changes, and every rate moves by 0.01 or less.

\paragraph{Choices the plan left open} (1) Incomplete raters: the rule above. (2) The low-rank-experience contrast counts a rater with 240 ratings as complete regardless of gates; three completed and all three passed. (3) A fit fails only if it errors or reports non-convergence; singular fits are kept. (4) The real-game accuracy test had no registered fallback; its low-tier fit gave a convergence warning, so we dropped the rater random effect, the primary test's first fallback. (5) The majority-vote fallback drops evenly split items, and (6) the robustness test's permutation fallback uses 100,000 permutations; neither fallback was needed. (7) The ranking test's tournament order is each strategy's mean composition-corrected label win probability over its tournament drafts. (8) The unfamiliarity covariate is standardized over each subset's analysis items. (9) Patch-2.57 exclusions use the 0.02 near-tie threshold.

\paragraph{All registered tests} Mixed-effects logistic models with crossed rater and item random effects (Laplace approximation, Wald intervals); the slider model is linear with a random slope per rater. Agreement with the model label on clear pairs: 0.50 (0.43--0.57); at near-tie thresholds 0.01 and 0.05, 0.52 and 0.49; with each single value function as the label, 0.50, 0.50, and 0.50; with the uncorrected label, 0.50. Anchored-vs-searching pairs: 0.33 (0.21--0.48). The unfamiliarity coefficients are $+$0.24 log-odds per SD (95\% CI $-$0.02 to 0.50) for maximum ensemble variance, $+$0.32 ($+$0.06 to 0.58) for mean variance, and $+$0.17 ($-$0.08 to 0.42) for matchup-level variance. All three lower interval bounds exceed the non-inferiority margin of $-$0.25 log-odds per SD. Real-game accuracy by tier: 0.60, 0.49, and 0.53 (Holm-adjusted $p$ = 0.004, 0.85, 0.85). Slider calibration slope 0.00 ($-$0.22 to 0.22). Low-rank-experience contrast: $-$0.09 log-odds ($-$0.55 to 0.36), one-sided $p = 0.66$. Ranking: Spearman $\rho = -0.19$ between the experts' Bradley--Terry strengths and the tournament order (exact one-sided $p = 0.71$ over all $10!$ orderings).

\paragraph{Model accuracy on the same real games (descriptive, post hoc)} Not in the registered plan. On the 570 anchor games, the model label (the mean of the three corrected value functions) picks the winner in 60.0\% (52.9--66.7), 54.2\% (47.1--61.1), and 55.8\% (48.7--62.7) of games at low, mid, and high tier, 56.7\% (52.6--60.7) overall; the enriched value function alone scores 64.7\%, 55.3\%, 55.8\%, and 58.6\% overall (Wilson 95\% intervals in parentheses, 190 games per tier). Pooled over raters, experts picked the winner in 58.7\%, 49.4\%, and 52.2\% of judgments, 53.4\% overall. With 190 games per tier the comparison is loose; above the lowest tier the experts are indistinguishable from chance and the models score 54--56\%, below the 57--58\% that draft-only models reach on held-out games.

\paragraph{Patch 2.57} Patch 2.57 went live on 2026-09-28, after every real game in the pool and before the machine drafts were generated by agents trained on earlier games. Every item and the instructions state the patch (2.55.17). Dropping items with any of the seven heroes whose balance changed in 2.57 leaves agreement at 0.49 (195 clear pairs), anchored-vs-searching agreement at 0.33 (97 items), and pooled real-game accuracy at 0.60, 0.45, and 0.51 by tier. A broader cut that also drops four heroes with gameplay bug fixes and two changed maps leaves 57 clear pairs (agreement 0.55) and too few anchored-vs-searching pairs (16) to estimate.